# Global monitoring of methane point sources using deep learning on hyperspectral radiance measurements from EMIT

Vishal V. Batchu*[1], Michelangelo Conserva*[1], Alex Wilson*[1], Anna M. Michalak[1,2], Varun Gulshan[1], Philip G. Brodrick[3], Andrew K. Thorpe[3], Christopher V. Arsdale[1]

[1]Google Research, [2]Carnegie Institution for Science, [3]Jet Propulsion Laboratory, California Institute of Technology

**Abstract**

Anthropogenic methane ($CH_4$) point sources are critical drivers of near-term climate forcing, safety hazards, and system-inefficiencies. Space-based imaging spectroscopy is an emerging tool for identifying emissions globally, but existing approaches largely rely on manual plume identification. Here, we present the Methane Analysis and Plume Localization with EMIT (MAPL-EMIT) model, an end-to-end vision transformer framework that leverages the complete radiance spectrum from the Earth Surface Mineral Dust Source Investigation (EMIT) instrument to jointly retrieve methane enhancements across all pixels within a scene. This approach brings together spectral information with spatial context to significantly lower detection limits. MAPL-EMIT simultaneously supports enhancement quantification, plume delineation, and source localization, even for overlapping plumes. The model was trained on 3.6 million physics-based synthetic plumes injected into global EMIT radiance data. Synthetic evaluation confirms the model's ability to identify plumes with high recall and precision and to capture weaker plumes relative to existing matched-filter approaches. On real-world benchmarks, MAPL-EMIT captures 84% of known hand-annotated NASA-L2B plume complexes across a test set of 1084 EMIT granules, while capturing roughly 1.5 times as many plausible plumes compared to human analysts. Further verification against coincident airborne data, top-emitting landfills, and controlled release experiments confirms the model's ability to identify previously uncaptured sources. By incorporating model-generated metrics such as spectral fit scores and estimated noise levels, the framework can limit false-positives. Overall, MAPL-EMIT enables high-throughput implementation on the full EMIT data catalog, shifting methane monitoring to a rapid, scalable paradigm for global plume mapping at the facility scale.

**Significance Statement**

Methane point-sources drive near-term climate forcing, yet tracking them globally remains challenging. We present a deep-learning framework that advances emission monitoring by directly analyzing satellite-based spectroscopic data and demonstrate its application to data from the Earth Surface Mineral Dust Source Investigation (EMIT) instrument. This approach provides the spatial context required to map gas concentrations across landscapes, significantly lowering detection limits and automating plume delineation and source localization, even for overlapping methane plumes from multiple sources. We validate the model against synthetic and real-world benchmarks, and demonstrate that it captures known emitters while identifying numerous previously undetected plumes. The automated end-to-end modeling pipeline enables application to the full EMIT archive, providing a compelling foundation for climate mitigation and facility-level accountability.

**Main Text**

**Introduction**

Atmospheric methane ($CH_4$) is a potent climate-forcing agent and the second largest contributor to anthropogenic greenhouse gas warming. With a 20-year global warming potential 81-86 times that of $CO_2$, $CH_4$ has driven approximately 25% of human-induced warming since the start of the industrial era. Given its high radiative efficiency and a relatively short nine-year atmospheric lifetime (Forster et al., 2021), reducing $CH_4$ emissions offers a critical "fast-action" pathway to mitigating global temperature rise. This urgency is reflected in the Global Methane Pledge (Ashfold et al., 2023), under which over 125 countries have committed to a 30% reduction in emissions by 2030.

Achieving these targets requires reliable monitoring of anthropogenic sources, which account for 60% of global $CH_4$ output. Within this 60%, emissions are concentrated within the waste (19%), agriculture (44%), and energy sectors (37%). The energy sector emissions consist of oil (33%), gas (25%), coal (28%) and bioenergy (14%) (International Energy Agency, 2025). Critically, much of the energy sector's footprint originates from localized point sources, making them primary targets for cost-effective mitigation.

Current space-based remote sensing of methane emissions faces a fundamental trade-off: instruments like the TROPOspheric Monitoring Instrument (TROPOMI) provide daily global coverage but at a coarse spatial resolution of 5.5 × 3.5 $km^2$ (Schuit et al., 2023) unsuitable for isolating individual sources. Conversely, targeted commercial imagers like GHGSat offer a high spatial resolution of 25 x 25 $m^2$ but very limited spatial coverage and restricted data access (Dowd et al., 2024). While multispectral sensors (e.g., Sentinel-2, Landsat) bridge this gap (Rouet-Leduc & Hulbert, 2024), they are limited to capturing very large emissions due to their coarse spectral resolution.

This study contributes to the growing literature that leverages observations from the Earth Surface Mineral Dust Source Investigation (EMIT) (Thorpe et al., 2023) to address this gap. Although originally designed for mineral mapping, EMIT's pushbroom hyperspectral imaging spectrometer balances these competing needs by providing high spectral and spatial resolution with substantial spatial coverage. Deployed on the International Space Station (ISS), EMIT captures 285 spectral bands (381 to 2,493 nm) in the visible through Short-Wave Infrared (SWIR) with a 60 m spatial resolution and a 80 km swath. This configuration provides the high Signal-to-Noise Ratio (SNR) and broad surface coverage necessary to characterize fine-scale methane absorption features across large regions. Existing approaches primarily rely on the Adaptive Matched Filter (AMF) (Thorpe et al., 2013, 2023; Thompson et al., 2015; Manolakis et al., 2014) for characterizing methane enhancements, followed by either expert-guided (e.g., Thorpe et al., 2023, Duren et al., 2025) or machine-learning-based (e.g. Ruzicka et al., 2023) plume segmentation. False positives caused by surface clutter, non-homogeneous terrain, and retrieval artifacts require careful examination, which presents challenges when scaling to global monitoring. Deep learning models built off of the underlying radiance data but which also include a spatial field via Convolutional Neural Networks (CNNs) or Vision Transformers, on the other hand, provide a potential scalable alternative (Coleman et al., 2026; Cusworth et al., 2019; Ehret et al., 2022; Rouet-Leduc & Hulbert, 2024; Vaughan et al., 2024). The core of the value added by these methods is the joint utilization of spectral characteristics and spatial patterning of those characteristics within the scene, which has the potential to suppress background noise and lower detection limits. The use of these methods is currently bottlenecked by a scarcity of diverse, high-quality annotated real-world data, however. While some studies have attempted to leverage synthetic plumes superimposed on real backgrounds (Gorroño et al., 2023; Rouet-Leduc & Hulbert, 2024), these efforts have largely been limited to multispectral sensors, small-scale datasets, or simplified radiative transfer models that struggle to generalize to complex operational environments.

Here we introduce the Methane Analysis and Plume Localization with EMIT (MAPL-EMIT) model, which directly utilizes the complete EMIT radiance spectrum in a vision transformer, providing spatial context to identify methane plumes across scenes. The model is trained on 3.6 million synthetic methane plumes superimposed on real EMIT backgrounds. The approach is used to automate enhancement quantification, plume segmentation and source localization, including for multiple overlapping plumes, unlocking the capability to rapidly investigate the full EMIT data record. We assess the model's ability to lower the detection limit for methane point sources. We further evaluate the model against a diverse set of real-world benchmarks, including plumes identified in the scientist-annotated NASA EMIT L2B product (Thorpe et al. 2026), an inventory of top-emitting landfills globally, coincident high-resolution airborne observations, and controlled release intercomparison studies.

## Results

### Evaluation on synthetic plumes

Evaluation against synthetic observations confirms the ability of MAPL-EMIT to detect and quantify methane enhancements, delineate plumes, and identify source location with high recall and precision.

For detection of individual plumes, the model precision increases from 0.80 for the weakest plumes (50-100 (kg/hr)/(m/s) which is equivalent to 125-250 kg/hr at typical wind speeds of ~2.5 m/s) to 0.97 for the most intense plumes (1600-3200 (kg/hr)/(m/s), indicating a decreasing rate of false positives as the plume intensity increases (Figure 1). The model recall ranges from 0.33 for the weakest plumes to 0.93 for the most intense ones, indicating an increasing ability to detect plumes at higher intensities, as expected.

The model also demonstrated high overall enhancement quantification accuracy on the synthetic test set. The normalized root mean squared error (RMSE) ranged from 6% for the highest intensity pixels (>2500 ppm-m) to 26% for pixels with enhancements in the range of 250-500 ppm-m, and greater for weaker enhancements (Figure S1). The symmetric mean absolute percentage error (SMAPE) followed a similar pattern (Figure S1). Plume-level integrated enhancement metrics (Figure S2) show that pixel-level errors often cancel out, leading to lower plume-level errors ranging from 8% for the highest emission plumes (1600-3200 (kg/hr)/(m/s)) to 24% for the weakest plumes (50-100 (kg/hr)/(m/s)), confirming strong quantification performance.

The model's ability to infer the source location was also strong, with a mean distance of 101m (1.68 pixels) between the true and estimated source location when averaged across all the different examined plume intensities.

The model is able to successfully identify and segment multiple overlapping plumes while estimating multiple source locations, with only modest impacts on plume detection (Figure 1) and enhancement quantification (Figure S1) for all but the weakest plumes. There was no impact on the ability of the model to identify the source location, with a mean distance between the true and estimated source location of 103m or 1.72 pixels when averaged across all examined plume intensities. An example of this is presented in (Figure S3). This is especially beneficial in industrial regions, where proximal emissions often cause plumes to coalesce downwind of emissions sources. Such overlapping plumes are not well captured by existing approaches.

The synthetic plume results suggest that MAPL-EMIT is able to capture weaker plumes relative to existing approaches, thereby presenting the potential to lower the detection limit of methane point sources. In the following sections we assess the degree to which these results generalize to real-world plumes.

### Performance on real-world benchmarks

We evaluate MAPL-EMIT against a wide range of data products to explore the model's ability to automatically detect and delineate real methane plumes and to localize sources. Because no reference set is complete, we rely on the combination of these comparisons to provide confidence in the model's performance. The comparisons include a comparison to EMIT L2B hand-annotated methane plume complexes (Thorpe et al. 2026), an examination of EMIT observations near the world's 25 top-emitting landfills (Zhang et al., 2025), an evaluation against coincident airborne observations from AVIRIS-3 (Coleman et al., 2026) and GAO (Ayasse et al., 2022), and validation against coincident controlled releases (Sherwin et al., 2024).

### ***NASA EMIT L2B plume complexes***

To validate detection capabilities against established benchmarks, we compare MAPL-EMIT's plume detections and plume boundaries against the EMIT L2B estimated methane plume complexes v002 (Thorpe et al., 2026; henceforth referred to as EMIT L2B plumes) across identical data granules. Model inference was run on 1084 granules, which contained 1689 EMIT L2B plumes. Because the EMIT L2B plumes have gone through multiple rounds of expert evaluation, the assumption is that these plumes contain relatively few false positives. Conversely, it is generally understood that many plumes are likely to be missed by this product due to background noise in the matched filter enhancements and due to a conscious decision to limit plumes to those with the highest confidence.

In order to focus on the higher confidence plumes identified by the MAPL-EMIT model, we filtered the model-identified plumes to those on-land (dropping any detections over water) and to those that appear in > 13 of 16 occurrences when processing overlapping tiles (i.e., strided inference), where the cluster size of 13 corresponds to the number of detections of a plume by the model over 16 strided inferences (as detailed in the Supplementary Information) (i.e., detection in > 81.25% of strided inferences). MAPL-EMIT initially identified 3672 plumes across the 1084 granules used in the comparison with the EMIT L2B plumes, and this number was reduced to 2210 after this filter was applied. An example tile from a granule is shown in Figure 2. Spectral fits confirm the presence of true plumes. Enhancement values between the methods disagree here, though there are many plausible explanations for this for a single plume, ranging from matched-filter sensitivities (Fahlen et al., 2024) to enhancement misestimation. The presence of high noise pixels in matched filter enhancements, including substantial negative values and anomalously high positive values, particularly near the plume source, dictate the visual scale. We do however see that across plumes overall, MAPL-EMIT enhancements align better with spectral fit enhancement estimates (see Discussion). The number of plumes for both models is broken out by the spectral fit enhancement in Figure 3. While MAPL-EMIT identifies more plumes across all intensities, the increase in the number of detections is significantly higher in the lower 0-100 ppm-m bins.

An evaluation of individual plumes shows that the final set of MAPL-EMIT plumes captures 84% of the EMIT L2B plume complexes. To prevent small boundary disagreements compromising statistics, MAPL-EMIT plumes included a 0.01° (~ 1.1 km, ~18 pixel) buffer, and a capture was considered successful if there was any plume intersection; results were very similar however with no buffer. This confirms the ability of the MAPL-EMIT model to automatically identify and delineate the high-intensity emissions detected by a matched-filter algorithm and expert-guided hand-annotation.

Conversely, only 67% of the MAPL-EMIT plumes were captured in the EMIT L2B plume collection (using the same buffering strategy as above), indicating that MAPL-EMIT identified twice as many plumes as human analysts did. This is likely due to a combination of three factors. First, MAPL-EMIT appears to be able to identify weak plumes that are difficult to differentiate from background noise in matched-filter-based approaches, and which are therefore not present in the EMIT collection (Figure S4). This presents the possibility of identifying many more real plumes relative to existing approaches. The second factor is that the EMIT L2B product may exclude high-intensity plumes due to stringent confidence thresholds in the review process or potential human omission during assessment (Figure S5). The third factor is that some of the MAPL-EMIT plumes are likely to be false positives, but the exact rate of false positives is challenging to assess given the lack of an absolute ground-truth. We examine this third factor next.

### ***False positive rate assessment in non-emission granules***

Quantifying false-positive (FP) rates in real-world satellite observations is challenging due to the absence of globally verified "zero-emission" ground-truth datasets. To address this, we evaluated MAPL-EMIT's performance on 20,000 granules where NASA JPL expects minimal plumes to be present. Per the EMIT L2B Greenhouse Gas ATBD (Brodrick et al., 2025), care is given to avoid false positives in plume delineation, and, as a result, some plumes are inevitably not included within this catalog. Also, human analysts may sometimes fail to identify some plumes. Therefore, while a reasonable indicator, this is not an exhaustive metric.

The model detected 1513 distinct plume instances across these 20,000 granules post filtering the model-identified plumes to those on-land and with an inference cluster size > 13 per plume. To differentiate plausible low-intensity emissions from likely false positives (such as retrieval artifacts or surface clutter), we followed the approach of Xiang et al. (2025) (see Methods “Spectral Fit Comparison”) to conduct spectral fit evaluation on all detections. This approach allows us to assess plumes and identify potential false positives. In doing so, we found that 355 of these detections are likely to represent genuine methane plumes.

The remaining 1158 unconfirmed detections, or approximately 0.07 plumes per 120 x 120 $km^2$ granule, provide a conservative estimate of the model's false-positive rate in complex real-world terrains. A majority of these plumes are on hilly terrain – both low and high albedo – or dense forests. By framing these unconfirmed detections as a conservative upper limit of the false positive rate, we demonstrate that the model maintains high precision while substantially expanding the envelope of detectable methane emissions.

***High-Emitting Landfills***
Landfills present a challenging retrieval environment for methane detection due to high surface heterogeneity and variable albedo. Landfills also often include multiple emission sources and/or spatially diffuse emissions, making methane enhancement detection and plume delineation difficult for existing algorithms. Zhang et al. (2025) had previously reported that the EMIT L2B methane complex product was able to detect potential plumes at 17 of the 25 top-emitting landfills globally.

Applying MAPL-EMIT at the locations of the 25 landfills demonstrated strong performance under these complex conditions. The model was able to identify potential plumes at 24 of the 25 landfills (examples in Figure S6). Importantly, the model had no information about underlying surface infrastructure or predominant wind direction and the correct identification of methane enhancements over these landfills is solely based on EMIT observations. In addition, MAPL-EMIT was often able to identify plumes for multiple EMIT overpasses, with the plume directions aligning with predominant wind directions at the time of the overpasses. An example of a site in Amman, Jordan, is shown in Figure S7. This location is a known persistent point source (Armstrong et al., 2024). MAPL-EMIT robustly captures persistent emission events across the full EMIT time series while avoiding false-positive detections in the surrounding area. Furthermore, the model resolves fine-grained plume morphology, capturing realistic, turbulent gas dispersion patterns. Here again, the inference was based solely on EMIT radiance information and wind information was not available to the model.

The application to landfills points to several additional strengths of MAPL-EMIT. First, as shown in Figure S6, the model demonstrates robust detection under partial cloud cover, a regime where traditional matched-filter methods often produce artifacts requiring the outputs to be masked out. Second, the fact that MAPL-EMIT is able to identify plumes, which originate at the landfill and are oriented along the predominant wind direction, for all but one of the top-emitting landfills, a substantially higher fraction than in the case of the EMIT L2B plumes (Figure S8), further supports the suggestions that MAPL-EMIT is able to successfully automatically detect weaker enhancements and delineate weaker plumes relative to existing approaches.

***Airborne Observations***
We leveraged airborne data with concurrent EMIT overpasses to further evaluate performance on real-world plumes. AVIRIS-3 served as the primary source for verification against airborne observations due to the availability of coincident overpasses (Coleman et al., 2026). We analyzed an EMIT granule overlapping with several AVIRIS-3 acquisitions (*AVIRIS-3 L2B Greenhouse Gas Enhancements, Facility Instrument Collection*, 2025), which identified five plumes. With a native spatial resolution of 0.3 m for this set of aerial collects, AVIRIS-3 was a robust verification source for lower-intensity emissions often obscured in spatially coarser satellite data.

We restricted the analysis to acquisitions collected within 30 minutes of the EMIT overpass. After resampling the aerial data to EMIT’s 60 m resolution, only three of the five original plumes remained discernible in the matched filter enhancements based on AVIRIS-3 data; the others became indistinguishable at the coarser scale (Figure 4 second panel). MAPL-EMIT successfully detected two of the three resolvable plumes without generating false positives (Figure 4 fourth panel) with an SNR of 8.60, higher than the SNR of resampled AVIRIS-3 matched filter which stood at 1.21. While the NASA L2B plumes also captured these plumes, a lower SNR of 0.40 in the standard matched-filters rendered them harder to isolate from background noise (Figure 4 third panel). SNR is computed as the plume enhancement mean minus background enhancement mean divided by the standard deviation of the background enhancement across the entire AVIRIS overpass (depicted in Figure 4, second panel). While EMIT's low resolution constrains morphological precision, MAPL-EMIT better replicates AVIRIS-3 plume geometry and orientation than EMIT L2B. This is most evident in the second plume, where L2B deviates significantly at the tail.

A comparison against airborne observations from the Global Airborne Observatory (GAO) (Ayasse et al., 2022) using the same 30-minute temporal constraint for coincident observations yielded similarly strong results (Figure S9).

***Controlled Releases***

We analyzed controlled methane release experiments with coincident EMIT overpasses (Reuland et al. 2026), which included five releases from Phase 2 and two releases from Phase 4 of the experiments. Phases 1 and 3 lacked unmasked EMIT overpasses.

MAPL-EMIT successfully detected five of the seven plumes (Figure S10), with a sixth plume showing up in the enhancements but not being identified as a plume due to not meeting the required threshold for plume instance probabilities (see Methods). Additionally, MAPL-EMIT did not produce any false positives at the location of release across any other EMIT overpasses of the locations. This detection rate confirms the model's ability to capture short-duration medium-intensity releases (~300-1000 (kg/hr)/(m/s)). Furthermore, spectral fit metrics across the detected plumes indicated strong agreement for most plumes (Figure S10, third column). Enhancements derived from these spectral fits (fit_enh) are also similar to enhancements predicted by the model (obs_enh) (Figure S10, fourth column).

## Discussion

### Challenges of methane emissions detection

Satellite methane detection faces four primary challenges. First, the lack of dedicated plume-detection instruments providing fully open radiance data. Most sensors that are presently used are not optimized for methane detection, being coarser either spatially (e.g., TROPOMI) or spectrally (e.g., EMIT, PRISMA, EnMAP) than would be ideal. Second, a severe scarcity of ground-truth data has historically impeded the development of benchmark datasets and objective intercomparisons for machine-learning-based algorithms. Third, the most established existing approaches rely on hand-annotation of plumes and evaluation by domain experts, making their rapid application across large data volumes challenging and costly. Finally, the high societal relevance of these emissions—involving economic losses, health risks, and significant climate forcing—creates a critical trade-off between missing real emissions (false negatives) and mistakenly identifying emissions (false positives). Because unreliable detections could threaten the credibility of remote sensing observations broadly, ensuring precision is essential to keeping satellite data actionable.

The MAPL-EMIT model presented here addresses these challenges by maximizing the information content extracted from EMIT observations. To circumvent the relatively coarse spectral resolution, the model directly utilizes all 285 EMIT spectral bands (except for a few as described in the Supplementary Information) and uses a vision transformer providing spatial context to jointly retrieve methane enhancement across all pixels within a scene. Analysis of patch embedding weights confirms that the model inherently prioritizes the three primary methane absorption manifolds (Figure S11). Sensitivity to other bands, such as those near the 700nm and 2000nm bands, may be attributed to water vapor or $CO_2$ absorption bands. Furthermore, distributed weights across the remaining spectrum likely facilitate robust background baselining and surface characterization. This comprehensive spectral utilization ensures robust discrimination of true plumes from surface-induced artifacts.

To resolve the historical lack of annotated ground truth, the model was developed using 3.6 million physics-based synthetic plumes superimposed on real EMIT scenes (see Methods). This approach allows for rigorous optimization and an explicit strategy for balancing detection trade-offs by adjusting plume mask thresholds (Figure S12). While we optimized thresholds to maximize the F1-score for the 100–200 (kg/hr)/(m/s) range, the framework is highly adaptable. As detailed in the Supplementary Information, we provide both model ensemble detection fractions (representing the number of plume detections across overlapping strided inferences) and spectral fit scores for each plume in the released data. This allows end users to dynamically adjust these parameters to suit their specific precision and recall requirements. Taken together, these innovations push the boundaries of emission detection by increasing sensitivity to smaller emissions and enabling the segmentation of overlapping plumes as described below.

### MAPL-EMIT model performance

Evaluating the model against a held-out synthetic test set demonstrates its ability to reliably capture plumes in the 100–200 (kg/hr)/(m/s) range, with F1 scores, precision, and recall all exceeding 50% for both individual and overlapping plumes. At a nominal wind speed of 2.5 m/s, this corresponds to leak rates of 250–500 kg/hr, representing an approximate 2-4x improvement over existing models (Tiemann et al., 2025, Ayasse et al., 2024), which put the 10% probability of detection at around 1000 kg/hr at 2.5 m/s windspeed. Furthermore, the model achieves a recall exceeding 80% for larger emissions above 800 (kg/hr)/(m/s) (Figure 1).

Comparisons with the EMIT L2B product confirm the ability to capture the majority (84%) of highly-vetted real-world plumes. Notably on the same set of 1084 granules included in the comparison to the EMIT L2B plumes, MAPL-EMIT achieved an enhancement SMAPE of 56% when comparing the model-estimated enhancement to the spectral fit enhancement across all the predicted plumes. For context, EMIT L2B plumes obtain an enhancement SMAPE of 117% on the same granules. This reduction in error demonstrates that the enhancements predicted by MAPL-EMIT correspond more closely with expected physical distributions, displaying fewer anomalous high and low-noisy enhancements.

In addition, MAPL-EMIT identifies hundreds of additional plumes in the same granules. While an investigation of all new detections is beyond the scope of this paper, a visual inspection of a subset of the additional identified plumes indicates that many are collocated with plausible methane leak locations (e.g., oil and gas infrastructure; Figure S5).

Retrievals near landfills and urban-industrial areas often suffer from false positives caused by surface albedo variability and background clutter. Traditional adaptive matched filters are highly sensitive to these artifacts, requiring conservative thresholds or manual review to maintain precision. In contrast, the model achieved a 96% recall across the world's top-emitting landfills while effectively suppressing surface-induced artifacts. Additionally, evaluations using synthetic artifacts (which mimic the methane spectral signature but lack realistic plume morphology) confirmed that the model effectively isolates true plumes from background noise (Figure S13). While hallucinations, i.e., model predicting fake plumes over these artifacts, can occur at high artifact intensities, the model generally discards these non-physical features.

In industrial regions, proximal leaks often coalesce downwind of emission sources. This overlap presents a significant challenge for traditional approaches. We find that MAPL-EMIT is able to disentangle these signals, enabling instance-level segmentation even in complex, multi-plume scenarios. Qualitative synthetic results in Figure S3 and comparisons to the EMIT L2B plumes (e.g., Figure 2) show successful plume delineation in the presence of overlapping plumes. Evaluation using synthetic plumes suggests that the performance of the model in terms of recall and precision is only marginally impacted by the presence of multiple plumes as shown in Figure 1.

### Opportunity for large-scale deployment

MAPL-EMIT provides an end-to-end deep learning framework for the automated detection and quantification of methane enhancements, plume delineation, and source localization from EMIT hyperspectral imagery. The model overcomes many of the limitations of traditional matched-filter algorithms, offering increased sensitivity to low-emission sources and a novel capability to resolve overlapping plume instances. Results on synthetic and real EMIT observations establish that transformer-based models trained on high-fidelity radiative transfer simulations can generalize effectively to global satellite observations. Integrated enhancement SMAPE statistics across varying emission rate bins (Figure S2) confirm the framework's capacity to accurately estimate integrated mass for emission rate quantification.

Taken together, this framework enables methane plume identification across the historical archive of EMIT observations. While performance on future data remains dependent on sensor calibration and degradation over time, a significant advantage of MAPL is its capacity for simple retraining, as it leverages synthetic data instead of relying on expert-annotated labels. The complete set of MAPL-EMIT outputs across the existing EMIT catalog has been made available, as detailed in the Data Availability section.

This plume database reveals a vast landscape of previously undocumented potential emissions, particularly at the lower end of the intensity spectrum, expanding the utility of global satellite archives to drive mitigation efforts. False positives remain an open challenge, with MAPL-EMIT, like other machine learning models in the literature, showing substantially more false positives than existing expert-vetted plume catalogues such as the EMIT L2B plume complexes. The false positive rate can be reduced by additional filtering based on spectral fit, estimated plume enhancement levels, and/or plume persistence levels as discussed in Results. To build a deeper understanding of the false positive landscape, subsequent research could further investigate scalable and automated methods for identifying false positives. Overall, MAPL-EMIT provides a valuable addition to global methane emissions monitoring by lowering the plausible detection limit in an operationalizable context, even if some post hoc review of model-identified plumes may still be required. Note also that the catalog is not expected to be complete, as indicated by the ~16% of the EMIT L2B plume complexes not captured by the model.

Future research will extend this synthesis pipeline to support multi-gas and multi-satellite retrieval and evolve the architecture toward end-to-end emission rate estimation, enabling direct leak rate emission without requiring post-hoc wind integration.

## Methods

MAPL-EMIT was designed to simultaneously delineate methane plumes, quantify their enhancements, and identify sources, directly from full spectrum radiance data. The model was trained on EMIT L1B radiances (Green, 2022) with injected synthetic methane plumes as outlined in Figure S14. Below we step through the process of synthetic plume generation, the specific model architecture, model training, evaluation on synthetic plumes and evaluation on real data.

### Dataset preparation

EMIT L1B at-sensor radiances (Green, 2022; Thompson, 2024) were utilized to train and evaluate the model. Using Google Earth Engine, we partitioned the data into non-overlapping 256 x 256 pixel tiles at the native 60m spatial resolution. To ensure high data quality, we filtered out tiles with more than 50% cloud cover or spacecraft obstructions, utilizing EMIT L2A surface reflectance masks. Additionally, to prevent the model from overfitting to specific surface or atmospheric conditions, we limited our sampling to a single, randomly selected timestamp for any given spatial location. The resulting dataset consisted of 235k tiles, which were divided into training (70%), validation (15%), and testing (15%) sets (containing 167k, 35k, and 33k tiles respectively). For synthetic plume validation and test analysis in the rest of the paper, we randomly subsampled 4096 tiles to optimize evaluation time. This subset provided a statistically significant sample of ~260 million pixels ($4096 \times 256 \times 256$) for robust performance characterization.

To ensure the physical validity of the radiative transfer, we exclude a small number of bands from the EMIT L1B radiance inputs prior to processing. Specifically, we remove seven bands around the order sorting filter, as well as the first three and last three bands to avoid noise, following the discussion in the EMIT Greenhouse Gases Algorithm Theoretical Basis Document (ATBD) (Brodrick et al., 2025) and the ISOFIT retrieval algorithm.

### Synthetic plume generation

We developed a large scale synthetic plume dataset using a Lagrangian puff model (Hurley, 1994) to simulate methane releases using the hyperparameters listed in Table S1. Plume origins were sampled randomly in clusters to ensure overlapping tails of plumes are present in some cases. Each source produces emissions with random intermittency similar to real releases. The model tracked the 2D movement of individual puffs dictated by a stochastic wind field. Simplex noise (Perlin, 2001) was used to simulate turbulence and eddies. Simulations were restricted to 2D because spaceborne instruments measure column-integrated enhancements; this approach significantly accelerated simulation speeds (see examples in Figure S15). Final concentrations and plume instance masks were derived by aggregating all puffs within the simulation region.

Simulations were performed on a $384 \times 384$ pixel grid where each pixel is 60m x 60m to match EMIT's native spatial resolution. To introduce variation, we generated plumes at $384 \times 384$ and during training, extracted random $256 \times 256$ pixel crops, while a deterministic central crop was used for evaluation and testing.

We generated 3.6 million plumes across 240k tiles, partitioned into training, validation, and testing sets consistent with the EMIT tiles. During training, EMIT tiles are randomly paired with plume tiles, varying by epoch. Evaluation and testing utilize fixed pairings for reproducibility. Each tile contains 15 plumes with corresponding instance masks and origin labels. We randomly select a subset of up to 10 plumes to inject during training for each tile. Every plume was initially simulated with a fixed emission rate of 1 mol/sec. This dataset has been publicly released (see Data Availability section for details) to facilitate further research. Subsequently, these plumes were scaled during radiative transfer to achieve target emission rates in the range of 100 to 10,000 kg/hr for training and evaluation. Plume masks were defined using a threshold of 0.00024 ppm-m (corresponds to 0.00001 millimoles/m$^2$) on the pre-scaling 1 mol/sec emission rate plumes and smoothed via erosion-dilation filtering to refine boundary edges.

Because there is a general lack of sufficient labeled data, it is not possible to comprehensively evaluate realism of the synthetic plumes against large quantities of real plumes. Instead, the realism (or lack thereof) of the synthetic data becomes evident when the trained model is used with actual EMIT data. If the synthetic data used to train the model had not been realistic, we would not have achieved useful results when applying the resulting model to real EMIT observations. In other words, the realism of the synthetic data is confirmed indirectly through the fact that the resulting model performs well on EMIT observations when evaluated against the real-world benchmarks.

### Plume Injection

Atmospheric gases exhibit distinct absorption spectra across varying wavelengths, a characteristic fundamental to $CH_4$ identification via hyperspectral data. Elevated $CH_4$ concentrations increase photon absorption at specific wavelengths. To model these interactions, we employed a line-by-line radiative transfer (Clough et al., 2005) approach using high-resolution spectral data from the HITRAN database (Gordon et al., 2022). Our calculations accounted for pressure-induced line broadening and thermal induced Doppler effects while incorporating continuum absorptions and emissions from $N_2$, $O_2$, and $H_2O$. This gives a similar transmittance profile at high spectral fidelity as is typically found in band models like MODTRAN (Berk et al., 2014). Additional details on the transmittance calculations are provided in The Supplementary Information.

For computational efficiency, and following the majority of matched-filter literature, we pre-calculated transmittance profiles into a Look-Up Table (LUT), spanning solar and instrument zenith angles as well as $CH_4$ concentration (following Thorpe et al., 2014).  Water vapor and aerosol variation were not considered in the LUT for simplicity, and we found that model performance was not substantially impacted by this choice (Figure S16). The LUT was then used to perform radiative transfer following the Beer-Lambert law (Swinehart, 1962). For simplicity, we ignore scattering effects, making the assumption that atmospheric path reflectance is small in the longer wavelengths where methane absorptions are strongest.

**Deep learning framework**

Our deep learning model takes EMIT radiances, angular metadata (solar and satellite zeniths), and crosstrack IDs as input to estimate methane enhancements. The ground truth labels consist of three components per plume: the methane enhancement (in millimoles/m²), a binary segmentation mask, and the plume origin.

Simultaneous resolution of all three of these components is a gap in the existing literature (Tiemann et al., 2025), largely because without simulation data it is very challenging to have accurate data at large enough scales to train a complex model. MAPL-EMIT's explicit leveraging of realistic, synthetic data helps overcome this challenge.

**Architecture**

The model utilizes a U-Net-like architecture (Ronneberger et al., 2015) (Figure S17) that couples a Swin-v2-S Transformer encoder (Liu et al., 2022) (~30M parameters) with a convolutional decoder. The encoder processes all the inputs mentioned above. It progressively downsamples input features, while the decoder reconstructs spatial resolution through a series of upsampling convolutional blocks. Skip connections bridge corresponding stages of the encoder and decoder, enabling the decoder to leverage multi-scale features, which is critical for precise spatial localization.

The final decoder feature map feeds into three parallel groups of prediction heads; each group consists of $P$ "slot" heads where $P$ corresponds to the maximum number of plumes the model can detect per tile, to disentangle individual plumes:

- Enhancement heads: Each head is trained to predict the methane enhancement for a single plume instance.
- Instance mask heads: These mirror the slot structure to generate a binary segmentation mask representing the spatial footprint of each plume.
- Origin heads: These mirror the slot structure to estimate plume origins, where each origin is represented as a binary circle of radius 15 pixels and treated as a binary segmentation task.

For this study, we set $P = 10$, which is sufficient for the typical plume density in the area covered by our $256 \times 256$ pixel tiles.

**Training**

Synthetic plumes were cropped to 256x256 pixels to add variation during training, ensuring that realistic scenarios such as diffused plume tails without visible origins are captured.  Plume enhancements were selected by randomly selecting a bin from [100, 200, 400, 800, 1600, 3200, 6400, 12800] ppm-m and then picking a value uniformly within that bin. These plumes were then injected into the 'background' EMIT radiance as outlined above. For evaluation, emission rates were log-uniformly sampled from 100 - 10,000 kg/hr to match a realistic real-world distribution of plumes (Jacob et al., 2022). Emission rates were converted to mol/sec = kg_per_hour / (3600 * (16.04 / 1000.0)), where 16.04 is the molar mass of methane, and plume enhancements were scaled linearly to achieve target intensities. 90% of training samples contained plumes as sampled above, but the remaining 10% contained no plumes to ensure that the model isn't biased towards always predicting a plume.

The model was trained using a multi-component loss function where enhancement losses are computed on a sublinear (square root) scale. This transformation allows the model to prioritize lower-intensity enhancements rather than being dominated by high-magnitude plumes. We ablated between linear, square root and log transformations (see Supplementary Information and Table S2), and square root performed the best. Establishing correspondence between unordered predictions and ground truth plumes requires solving an assignment problem, which we addressed using the Hungarian matching algorithm (Stewart et al., 2016).

$$\mathcal{L}_\delta(y, \hat{y}) = \begin{cases} \frac{1}{2}(y - \hat{y})^2 & \text{for } |y - \hat{y}| \le \delta, \\ \delta(|y - \hat{y}| - \frac{1}{2}\delta) & \text{otherwise.} \end{cases} \quad (1)$$

For a single prediction-ground truth pair $(i, j)$, the enhancement loss is the Huber loss ($\mathcal{L}_\delta$) (Huber, 1964) (Equation 1) on square-rooted labels ($y$) and predictions ($\hat{y}$) where $\delta$ is the critical threshold, while both the mask and origin losses utilize the binary cross-entropy (BCE) loss. The Hungarian algorithm minimizes the total cost across matches to find the paired slot loss, $\mathcal{L}_{slot}$ Additionally, we applied auxiliary "total enhancement", "total semantic" and “total origin semantic” losses to guide the model toward conserving total methane mass.

The final training objective, $\mathcal{L}_{\text{final}}$, is a weighted sum of the paired slot loss and the auxiliary total enhancement loss.

Let $\hat{\sigma}$ be the optimal one-to-one matching (permutation) found by the Hungarian algorithm. The final loss is defined as:

$$\mathcal{L}_{\text{final}} = \lambda_{\text{slot}}\mathcal{L}_{\text{slot}} + \lambda_{\text{total}}\mathcal{L}_{\text{total}} \quad (2)$$

Where $\mathcal{L}_{slot}$ is the minimum sum of paired losses across $P$ slots defined as:

$$\mathcal{L}_{\text{slot}} = \sum_{i=1}^{P} \left( \lambda_{\text{enh}}\mathcal{L}_{\text{huber}}(\sqrt{E_i}, \sqrt{E^{\text{gt}}_{\hat{\sigma}(i)}}) + \lambda_{\text{mask}}\mathcal{L}_{\text{bce}}(M_i, M^{\text{gt}}_{\hat{\sigma}(i)}) + \lambda_{\text{origin}}\mathcal{L}_{\text{bce}}(O_i, O^{\text{gt}}_{\hat{\sigma}(i)}) \right) \quad (3)$$

The second component $\mathcal{L}_{\text{total}}$ constrains the aggregated sums for enhancements and the max of instance masks for plume masks ($M$), and origins ($O$):

$$\mathcal{L}_{\text{total}} = \lambda_{\text{e_total}}\mathcal{L}_{\text{huber}} \left( \sqrt{\sum E_i}, \sqrt{\sum E^{\text{gt}}_i} \right) + \lambda_{\text{m_total}}\mathcal{L}_{\text{bce}} \left( \max M_i, \max M^{\text{gt}}_i \right) + \lambda_{\text{o_total}}\mathcal{L}_{\text{bce}} \left( \max O_i, \max O^{\text{gt}}_i \right) \quad (4)$$

We selected the weights for each of the loss components to ensure that the scale of all the slot level losses are ~2x the scale of all the top level aggregated losses (Table S3). Additionally, at each slot/top level, we ensure that the losses across heads are roughly at the same scale. We do this after the model stabilizes at ~50k steps. The final loss weights were set as follows: $\lambda_{e\ total} = 6.0$, $\lambda_{m\ total} = 1.0$, $\lambda_{o\ total} = 19.0$, $\lambda_{enh} = 2.6$, $\lambda_{mask} = 1.0$, and $\lambda_{origin} = 24.0$.

Lastly, we also upweighted plume pixels by 30.0 on the enhancement losses and 1.0 (no upweight) on the plume mask and origin losses (across both the slot loss and the total loss) to ensure that the model focuses more on plume pixels compared to the background pixels. An upweight sweep ablation is reported in the Supplementary Information and Table S4.

The model was implemented in Jeo (Team, 2025) and trained using the AdamW optimizer (Loshchilov & Hutter, 2017) with parameters $\beta_1$=0.9, $\beta_2$=0.999, eps=1e-08 and a weight decay of $1e^{-9}$. We employed a cosine decay learning rate schedule with a linear warmup for the first 1000 steps to a peak learning rate of $3e^{-4}$, followed by a decay to a minimum of 0. The training consisted of two stages, the first using a batch size of 128 for 500k steps and the second using a batch size of 256 for 1M steps fine tuning the model from the first stage. All experiments were conducted on Google TPUs (32 chips), with the complete training process taking approximately 96 hours.

## Model Inference

To transition MAPL-EMIT from fixed-window patch predictions ($256 \times 256$ pixels, ~234 $km^2$) to EMIT granule level predictions (~5,700 $km^2$), we developed a large-scale inference pipeline that operates at a granule level designed to produce two distinct outputs: a methane enhancement raster and a set of plume instances, including individual plume enhancements.

To generate the enhancement raster, the pipeline processed full granules using overlapping spatial strides of 64 pixels. Multi-slot predicted enhancements were summed and reconstructed using a Hanning window function (Harris, 1978), which seamlessly merged overlapping tiles into a continuous raster by weighting central pixel predictions to eliminate edge discontinuities. To produce the plume instance collection, the pipeline resolved redundant detections caused by the overlapping strides through a custom spatial clustering algorithm inspired by DBSCAN (Ester et al., 1996). This process clustered candidate plumes into distinct entities by jointly evaluating spatial proximity of predicted plume origins and plume-head enhancement correlation. Finally, all isolated candidates were subjected to the physics-based spectral vetting pipeline (see “Spectral fit comparison” section below) to produce spectral fit metrics. A description of the entire pipeline is provided in the Supplementary Information.

## Model Evaluation

### *Synthetic evaluation*

We evaluated the model's tile-level performance across three core tasks: enhancement quantification, plume detection, and origin estimation. A systematic evaluation on a held-out synthetic test set was essential, as it provided precise ground-truth labels often unavailable in real-world observations. By stratifying these results across simulated wind and emission conditions, we quantified sensitivity limits and established operating thresholds for large-scale monitoring.

Performance was assessed using distinct metrics for each task:

1. Enhancement Quantification: We computed label normalized Root Mean Squared Error (RMSE) (Guanter et al., 2021; Rouet-Leduc & Hulbert, 2024) and Symmetric Mean Absolute Percentage Error (SMAPE) (Makridakis, 1993). These metrics were averaged across all $P$ output slots to get the final numbers.
2. Plume Delineation: We calculated precision, recall, F1 score, and integrated enhancement SMAPE across plume instances. Precision-Recall (PR) curves were generated to identify the optimal operating threshold based on predicted mask probabilities.
3. Source Localization: We computed the precision, recall and mean origin euclidean distance for true positives in meters.

All enhancement metrics were stratified by grouping pixels into buckets of label enhancements [0, 5, 25, 50, 125, 250, 500, 1000, 2500, 25,000] ppm-m. Plume detection metrics were stratified by grouping ground-truth plumes into buckets of emission rate divided by wind speed (ERBWS) [50, 100, 200, 400, 800, 1600, 3200], which correspond to approximately [125, 250, 500, 1000, 2000, 4000, 8000] kg/hr at $2.5\ m/s$ wind speeds. We evaluated ERBWS bins because emissions and wind speeds exhibit an inverse linear sensitivity relationship, facilitating consistent performance tracking across distinct categories. The validation and test splits utilized a log-uniform emission rate distribution because it gives us a roughly similar number of plumes in each logarithmic ERBWS bucket during evaluation.

We selected the final model, operating plume and origin probability thresholds by maximizing the plume instance F1 score for the 100–200 ERBWS bucket on the validation split (note that in Figure S12, 0.3 is the optimal threshold that maximizes the F1 score but all our parameter selection was done on the validation split to avoid overfitting, where 0.4 turned out to be the optimal threshold). This range ($250–500\ kg/hr$ at $2.5\ m/s$) represents our primary focus, as it encompasses a significant fraction of real-world methane leaks following a heavy-tailed distribution (Duren et al., 2019; Jacob et al., 2022).

To derive discrete predicted origins from the model's origin probability maps, we utilized a threshold-based weighted centroid approach. For each plume slot, we identify all pixels where the predicted probability exceeds a threshold of 0.3 (this threshold gave us the best metrics on the synthetic validation split). The final estimated origin coordinate (x,y) is computed as the weighted mean of these pixel locations, using the predicted probabilities $P_i$ as the weights (Eqn. 5).

$$\hat{x} = \frac{\sum P_i x_i}{\sum P_i}, \quad \hat{y} = \frac{\sum P_i y_i}{\sum P_i} \quad (5)$$

All the metrics are calculated after aligning predicted plumes, origins, and enhancements with ground-truth labels via Hungarian matching of semantic masks. For the plume detection metrics, a prediction was classified as a True Positive (TP) if its IoU with a ground-truth mask exceeded 0.25. Predictions failing this threshold were counted as False Positives (FP), while unmatched ground-truth plumes were False Negatives (FN). Similarly, for origin estimation, a prediction was classified as a True Positive if it was within 600m in terms of euclidean distance. The model was trained to produce empty outputs for detection and origin slots that do not correspond to a ground-truth plume.

Figure S12 illustrates performance trade-offs based on the plume probability threshold, depicting the PR curve for the 100–200 ERBWS bucket, the balance between true positives and false positives, and the resulting precision and recall curves. These curves demonstrate the model's ability to maintain high precision via a plume probability threshold adjustment, which is critical for operational monitoring where false alarms must be minimized.

***Spectral fit comparison***

To verify the model's real-world performance, we employed spectral fit comparisons (Xiang et al., 2025) as an independent verification metric across all real-world evaluations. This method serves to verify the physical validity of detected plumes by comparing the observed radiance against modeled spectra (Figures S18, S19, S20).

As described in (Xiang et al., 2025), the methodology proceeds as follows. We define in-plume pixels by taking the model predicted mask, excluding the top 1% of model predicted enhancements, and sampling 3x3 regions around the top 30 remaining pixels. Due to spatial overlap, this typically yields 100–150 unique pixels. To select out-of-plume (background) pixels, we expand the plume mask by 200 pixels, remove any other model predicted plumes in the granule, and exclude pixels with enhancements > 30 ppm-m. We then use Hungarian matching on non methane absorption bands to pair each in-plume pixel with its closest background match. After computing mean radiance ratios for the in-plume and out-of-plume pixels individually, we apply a least-squares optimization to fit parameters, including a fitted enhancement (fit_enh). We then compare the mean of the model's predicted enhancements across the in-plume pixels (obs_enh), against fit_enh. Spectral fit scores are assessed using correlation error ($D_{cor}$) and normalized error ($D_{norm}$) obtained by comparing the fitted and observed transmittance ratios.

A correlation error ($D_{cor}$) below 0.4 and a normalized error ($D_{norm}$) below 0.5 were found by (Xiang et al., 2025) to typically identify valid methane plumes. While high error values of $D_{norm}$ above 0.8 generally suggest false positives, detections with elevated $D_{cor}$ or $D_{norm}$ were still considered legitimate if they coincided with high spectral fit estimated methane enhancements (fit_enh) exceeding 100 ppm-m.

This spectral analysis is particularly critical for low-intensity plumes that lack traditional ground-truth labels. We therefore use these thresholds to identify potential true plumes in the 20,000 granules used in the false positive rate assessment and to evaluate the additional plumes identified in the granules used in the comparison to the NASA EMIT L2B plume complexes.

***Real world evaluation***

To evaluate the model's generalizability from synthetic training to real-world observations, we compiled a diverse suite of verification and validation datasets. While the inherent scarcity of high-resolution ground-truth data often constrains rigorous quantitative assessments of sensitivity and precision, we leveraged a diversity of available observational benchmarks to ensure a robust performance evaluation. We compare against the EMIT L2B plume complexes as described in the main results, and perform several additional checks as outlined below.

***High resolution aerial observations***

To evaluate model performance against high-fidelity aerial sources, we utilized observations from the Airborne Visible/Infrared Imaging Spectrometer-3 (AVIRIS-3) (Chlus et al., 2025) and the Global Airborne Observatory (GAO) (Ayasse et al., 2022). These sensors provide superior spatial resolution and SNR compared to spaceborne instruments. We specifically leveraged methane enhancements from these aerial datasets in regions with coincident EMIT overpasses. By comparing MAPL-EMIT's predicted plumes and enhancements against the L2B methane products from AVIRIS-3 and GAO, we established a benchmark for sensitivity and false-positive suppression in real-world environments.

***Controlled release experiments***

To validate the model's detection sensitivity and enhancement quantification accuracy, we utilized data from the Stanford controlled methane release experiments conducted August 1 2024 to December 31 2025 (Reuland et al., 2026). These experiments were conducted near a dedicated field site in Arizona ($32.813^{\circ}, -111.796^{\circ}$). We specifically analyzed five releases from the Stanford Phase 2 and two releases from the Phase 4 campaigns, allowing for a rigorous assessment of the model's ability to recover plumes from low-intensity, short-duration releases.

***Top-emitting landfills***

To validate the model’s performance in complex environments, we evaluated its recall across 25 of the highest-emitting municipal solid waste landfills globally (sampling one random EMIT timestamp at each location). These sites were selected based on a recent hyperspectral survey that identified major methane super-emitters using EMIT and other high-resolution sensors (Zhang et al., 2025). Landfills represent a challenging retrieval domain due to high surface heterogeneity and variable albedo, which often induce artifacts in traditional physics-based algorithms. By benchmarking against these established emitters, we assessed the model’s ability to maintain high detection sensitivity and suppress false positives in terrains where background clutter typically limits automated monitoring.

**Acknowledgments**

We wish to express our gratitude to (listed alphabetically) Burak Ekim, Carl Elkin, Tal Geller, Omry Gillon, Nita Goyal, Mansi Kansal, Roy Nadler, John Platt, Sergei Shames, Bijoy Shetty, Aaron Sonabend, Deepika Sukhija, and Shahar Timnat for their insightful contributions and discussions regarding the methane retrieval framework. We thank Maxim Neumann and Anton Raichuk for their technical support with the Jeo library. Additionally, we'd like to thank Frances Reuland and Adam R. Brandt for working with us to assess model performance on the Stanford controlled-releases. We are also grateful to David R. Thompson and Robert O. Green at NASA JPL for their assistance in the real-world verification of MAPL-EMIT.

A portion of this research was carried out at the Jet Propulsion Laboratory, California Institute of Technology, under a contract with the National Aeronautics and Space Administration (80NM0018D0004). US Government Support Acknowledged.

During the preparation of this work the authors used Google Gemini-3 in order to polish the text of the initial draft of the paper. After using this tool/service, the authors reviewed and edited the content as needed and take full responsibility for the content of the published article.

**Data availability**

The datasets used during this study are made available to support reproducibility and further research. The dataset from NASA (EMIT L1B, L2A, L2B CH4 ENH, L2B CH4 PLM, and AVIRIS-3) are available from the NASA catalog at https://www.earthdata.nasa.gov/data/catalog. The Global Airborne Observatory (GAO) data can be accessed at https://data.carbonmapper.org.

To facilitate public access and ensure reproducible research, we will release the synthetic plumes that were used to train our model and the fully trained model on the Kaggle platform with the supporting code on GitHub upon final publication.

To enable large-scale analysis, we will release the model inference outputs across the full EMIT data record as Google Earth Engine image collections, comprising per-granule total enhancement rasters and per-plume instance collections with segmented methane enhancements for every detected source upon final publication.

## Figures

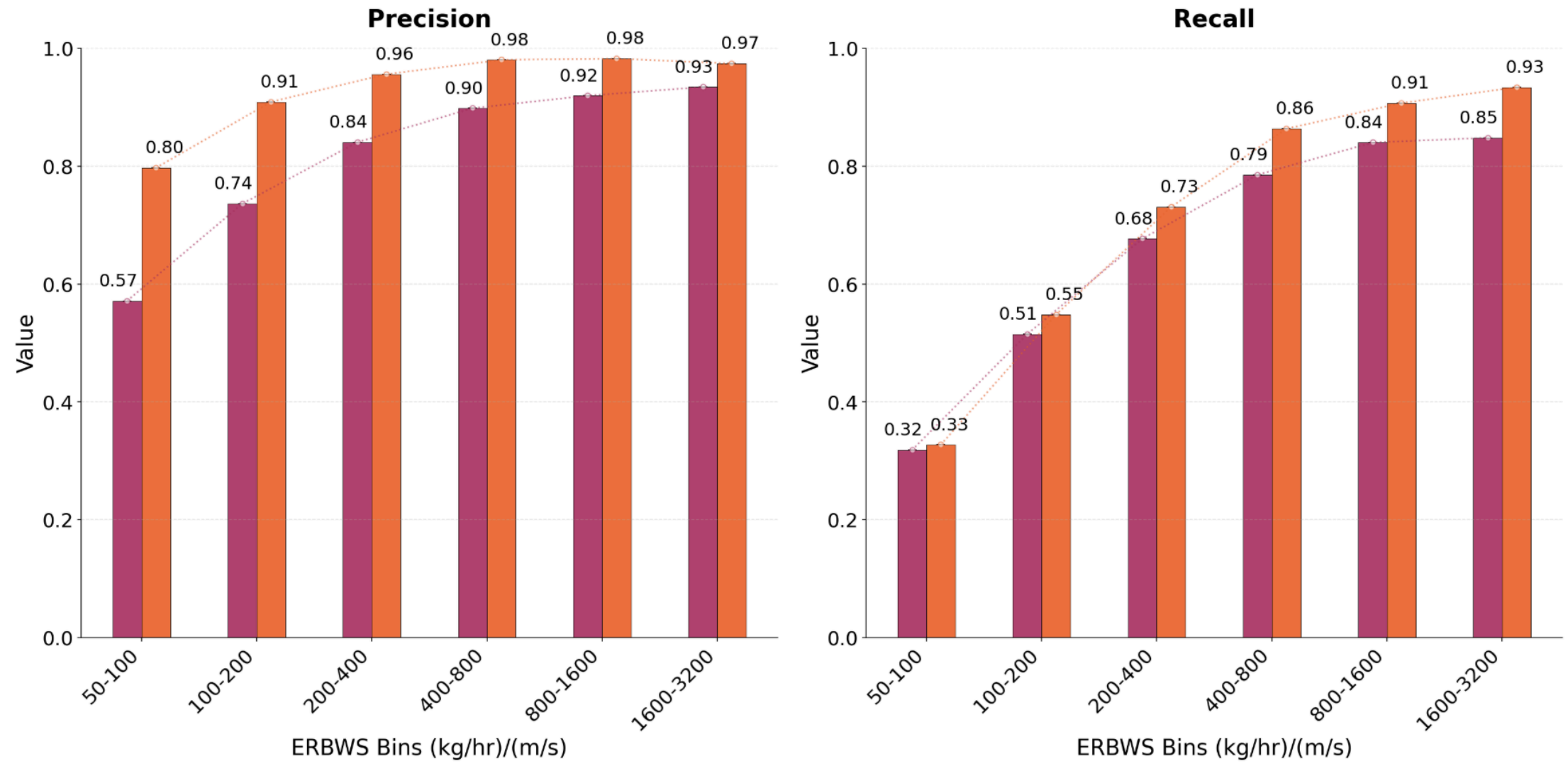

**Figure 1.** MAPL-EMIT exhibits high precision and recall across plumes with a wide range of emission rates on the synthetic plume test split. Precision (left panel) and recall (right panel) are binned for different plume intensity levels, defined as emission rate (kg/hr) divided by 10m elevation wind speed (m/s) (ERBWS), for scenes with single plumes (orange) as well as for multiple overlapping plumes (in the same EMIT tile) (magenta). The numbers above the bars indicate the precision and recall for each ERBWS bin.

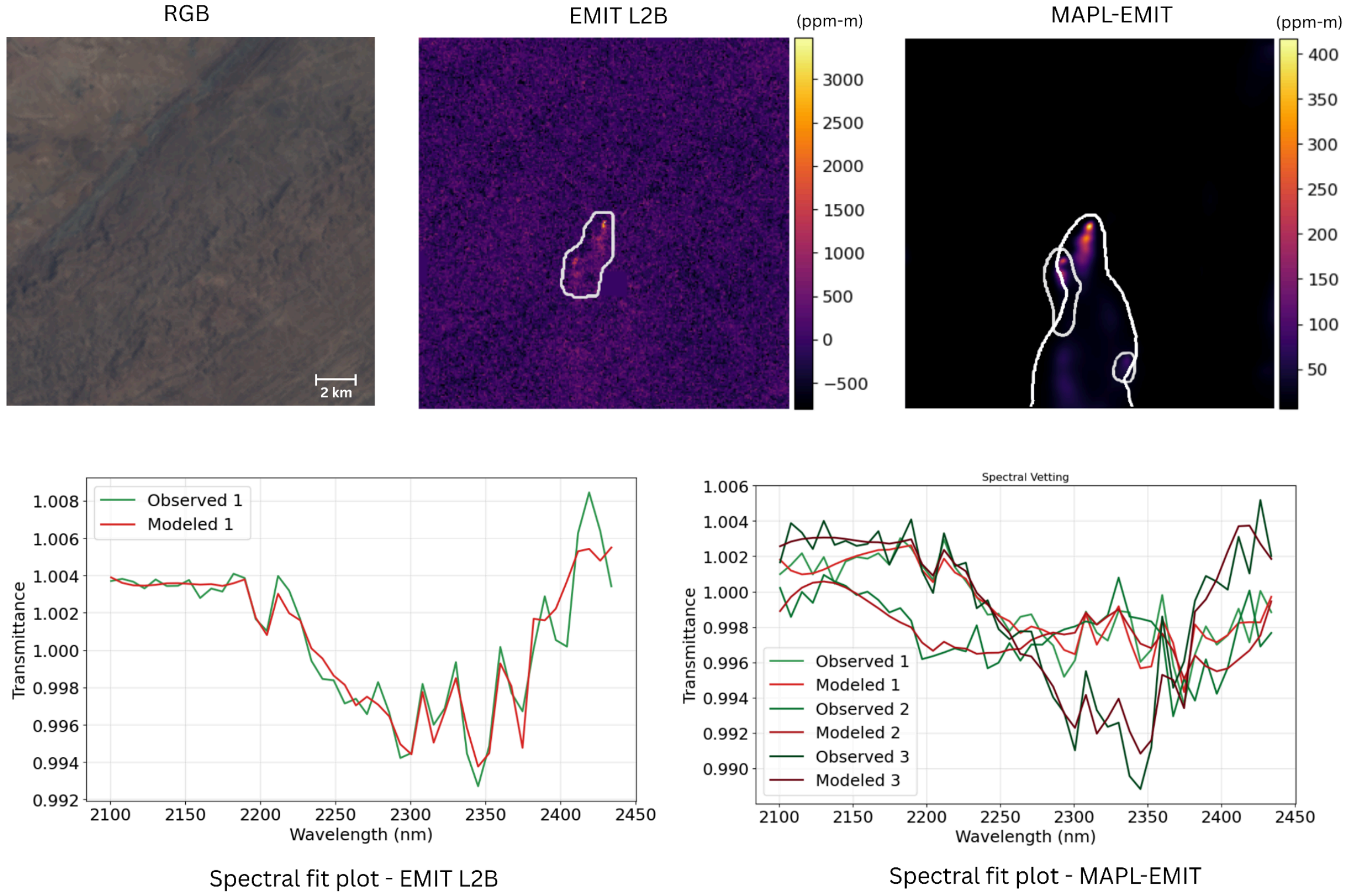

**Figure 2.** MAPL-EMIT detection and disambiguation of overlapping methane plumes. Top row: RGB (left), NASA L2B enhancements with plume masks (middle), and MAPL-EMIT enhancements with plume delineation (right). Bottom row: Spectral fits provide physical verification. Observed transmittance ratios align with modeled methane absorption manifolds, supporting plume authenticity. While the first MAPL-EMIT plume shows a robust spectral fit, the second appears less definitive, likely due to its weaker intensity.

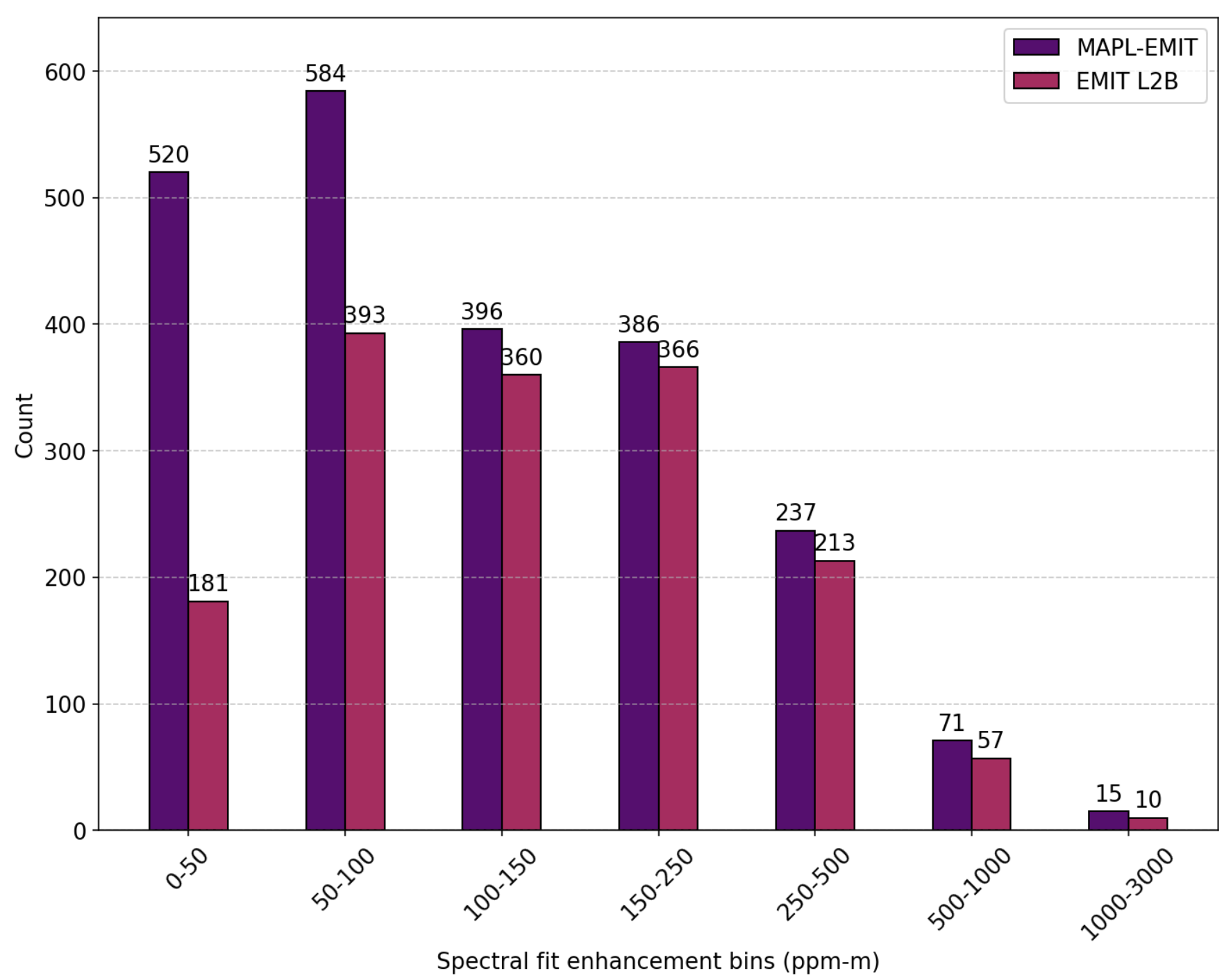

**Figure 3.** MAPL-EMIT identifies more potential plumes than the official EMIT L2B plume complexes. Plumes are binned by spectral fit enhancement (see Methods). MAPL-EMIT (purple) consistently identifies more plumes across all intensities, specifically more than 2x the number of detections for the 0-50 ppm-m bin. The numbers above the bars indicate the number of identified plumes in each enhancement bin.

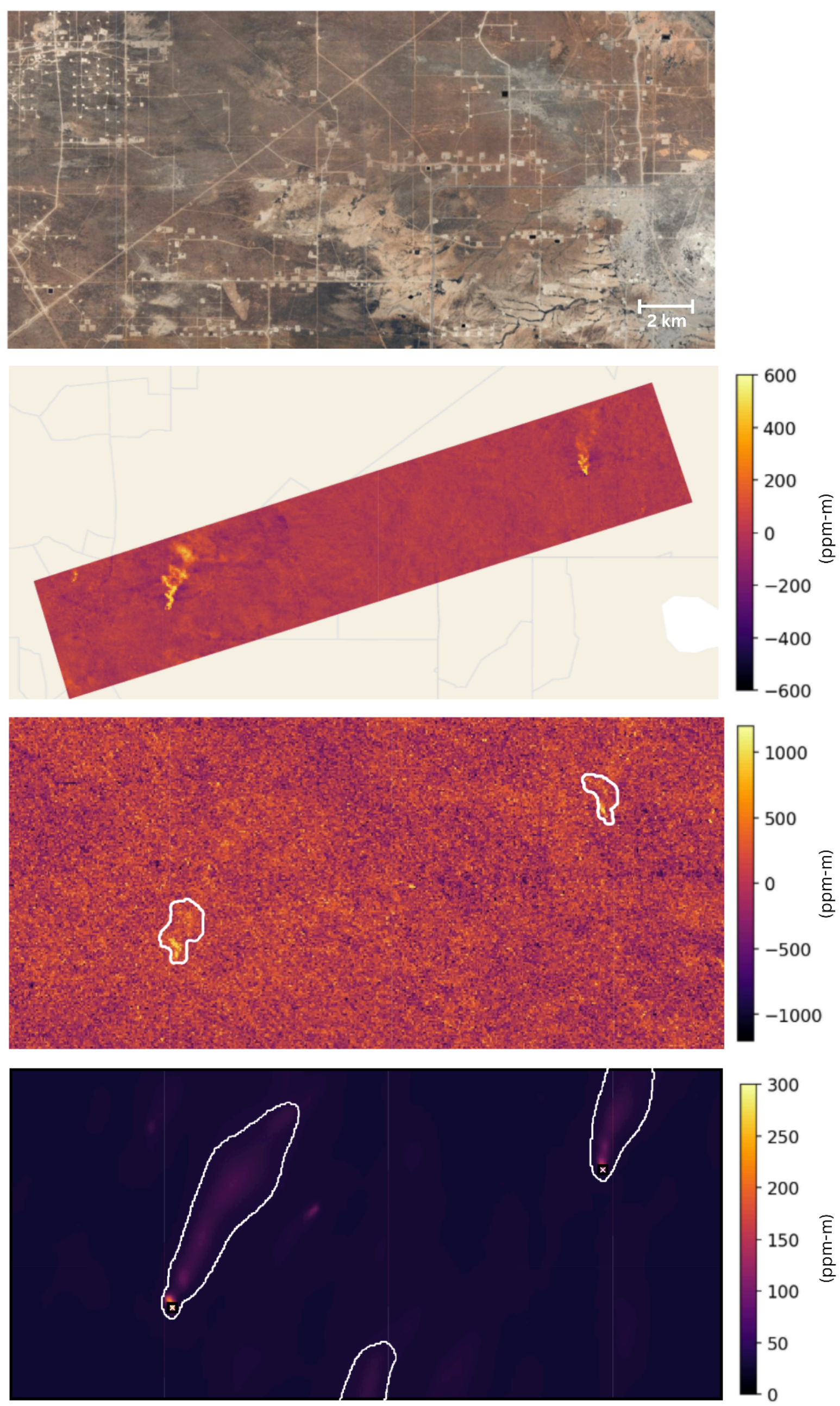

**Figure 4.** Enhanced signal-to-noise ratio and verification of MAPL-EMIT plumes against high-resolution aerial enhancements from AVIRIS-3. The panels display (from top to bottom): RGB, coincident AVIRIS-3 airborne observations resampled to 60m resolution, NASA L2B enhancement with identified plume complexes on top, where plumes are hard to distinguish from background noise, and MAPL-EMIT enhancements and plumes, which successfully suppress surface artifacts and accurately predict plume morphologies aligned with AVIRIS-3 observations.

## Supporting Information for

# Global monitoring of methane point sources using deep learning on hyperspectral radiance measurements from EMIT

Vishal V. Batchu*[1], Michelangelo Conserva*[1], Alex Wilson*[1], Varun Gulshan[1], Anna M. Michalak[1,2], Philip G. Brodrick[3], Andrew K. Thorpe[3], Christopher V. Arsdale[1]

[1]Google Research, [2]Carnegie Institution for Science, [3]Jet Propulsion Laboratory, California Institute of Technology

### Synthetic Data Generation

**Transmittance Calculations.** We acquired atmospheric transmittances at a spectral resolution of 0.00025 µm over a range of 0.38 to 2.5 µm for a US standard atmosphere (U.S. Standard Atmosphere, 1976, 1976). These data were modeled using SpectralCalc (Gordley et al., 1994) with a nadir-viewing geometry and an observer height of 600 km, hereafter referred to as "standard atmosphere transmittances."

A synthetic atmosphere was subsequently generated by modifying the standard US atmosphere to include a localized methane enhancement. We added 1 ppm of methane from the Earth's surface through the initial 100m of the atmospheric column, followed by a linear decay to 0 ppm over the subsequent 100m. This configuration represents a constant vertical profile of methane release used across all simulations, contributing a total vertical column concentration of 150 ppm-m visualized in Figure S21. While sampling varied methane addition profiles would be ideal, we selected this representative atmosphere to maintain computational scalability. Because the total methane mass added remained fixed, minor variations in the vertical distribution result in negligible differences in the column-integrated absorption signal measured by the instrument. Transmittances were then obtained for this modified atmosphere, hereafter termed "methane transmittances."

**Spectral Responses.** Laboratory-calibrated spectral response functions (SRFs) for the EMIT bands are accurately approximated by Gaussian distributions (Thompson et al., 2024). For modeling simplicity, we treated each band's SRF as a Gaussian centered at the band mean wavelength. The standard deviation was derived directly from the measured Full Width at Half Maximum for each respective band.

**Lookup table based on Beer-Lambert absorption.** The transmittance of an atmosphere containing *x* ppm-m of added methane at a given wavelength (λ) is derived by linearizing the Beer-Lambert law. This formulation assumes that the differential absorption remains non-saturating within the modeled concentration range. Because the methane levels utilized in this study did not lead to significant saturation of the absorption lines, the added concentration was applied directly as an exponent (Eqn S1).

$$t^{x}_{\mathrm{methane}}(\lambda) = t_{\mathrm{std}}(\lambda) \cdot \left( \frac{t^{150}_{\mathrm{methane}}(\lambda)}{t_{\mathrm{std}}(\lambda)} \right)^{\frac{x}{150}} \quad \text{(S1)}$$

To account for the total photon path, we adjusted the transmittance by a path length multiplier determined by the solar and satellite zenith angles (Eqn S2). Since SpectralCalc data assumed a single vertical pass of the atmosphere, we performed this adjustment using the same exponential scaling logic (Equation S3) for both $CH_4$ and standard atmosphere transmittances ($t_{std}$) as well.

$$\mathrm{path_length_multiplier} = \frac{1}{\cos(\theta_{\mathrm{solar}})} + \frac{1}{\cos(\theta_{\mathrm{satellite}})} \quad \text{(S2)}$$

$$t_{\text{methane}}^{x,\text{path_length_multiplier}}(\lambda) = t_{\text{std}}(\lambda) \cdot \left( \frac{t_{\text{methane}}^{150}(\lambda)}{t_{\text{std}}(\lambda)} \right)^{\frac{x \cdot \text{path_length_multiplier}}{150}} \quad \text{(S3)}$$

We populated a Look-Up Table (LUT) with transmittance ratio multipliers for all combinations of methane concentrations and path length multipliers. This pre-computation was executed once due to its high spectral resolution and computational cost (Eqns S4, S5 and S6).

$$\text{methane_concentrations} = \{0, 100, \ldots, 900\} \cup \{1000, 2000, \ldots, 9000\} \cup \{100000\} \quad \text{(S4)}$$

$$\text{path_length_multipliers} = \{1, 2, 3, 4, 5, 100\} \quad \text{(S5)}$$

$$\begin{aligned}
&\textbf{for each } conc \in \text{methane_concentrations}: \\
&\quad \textbf{for each } \text{path_multiplier} \in \text{path_length_multipliers}: \\
&\qquad LUT(conc, path_length_multiplier) = \log \frac{\sum_{\lambda \in \mathcal{L}} \left[ \text{SRF}(\lambda) \cdot \left( t_{\text{methane}}^{x}(\lambda) \right)^{path_length_multiplier} \right]}{\sum_{\lambda \in \mathcal{L}} \left[ \text{SRF}(\lambda) \cdot \left( t_{\text{std}}(\lambda) \right)^{path_length_multiplier} \right]}
\end{aligned} \quad \text{(S6)}$$

During data generation, bounding entries were identified via a LUT search, followed by bilinear interpolation in log-space (Eqn S7). Log-space interpolation was essential to maintaining accuracy because concentration and path length factors remain exponential even after spectral band averaging. The final synthetic radiance was then derived (Eqn S8) using the multiplicative factor computed in Eqn S7.

$$R_{\text{final}}(\text{conc}, \text{path_length_multiplier}) = \exp(\text{bilinearInterp}(\log(\text{LUT}))) \quad \text{(S7)}$$

$$L_{\text{syn}}(x) = L_{\text{std}} \cdot R_{\text{final}}(conc, \text{path_length_multiplier}) \quad \text{(S8)}$$

Our radiative transfer assumed a constant methane concentration along the photon path. While an oblique viewing geometry (maximal angles of ~45 with methane altitudes up to 200 m) result in a path integral spanning approximately 3.33 pixels, we did not apply path-integrated mixing as the majority of our synthetic plumes exceeded this 3.33 pixel scale and this only introduced a marginal inaccuracy.

To simplify the radiative transfer model, aerosols and water vapor were excluded from the retrieval framework. Sensitivity analyses in Figure S16 where the SMAPE of the integrated enhancements between model detected plumes and NASA plumes indicate a lack of significant correlation between retrieval error (SMAPE) and either water vapor concentration or AOD 550. This analysis was conducted on all plumes detected by MAPL-EMIT that also appear in the NASA L2B plume complexes catalog across the subset of tiles used in the intercomparison (see Main Text). For these instances, we used the NASA L2B plume mask as the ground-truth label for the integration process. The stability of the binned medians (and 25, 75 percentile ranges) across these atmospheric variables suggests that the model is robust to these factors within the observed ranges, justifying their omission to improve computational efficiency.

**Input normalization.** We implement a two-stage normalization pipeline to standardize EMIT L1B radiances prior to passing them as inputs to the model.

1. We perform a solar spectrum normalization to remove the dominant spectral shape of the solar irradiance (Planck, n.d.). We compute the theoretical spectral radiance of the Sun, modeled as a blackbody at T=5778 K using Planck's Law. This high-resolution blackbody curve is convolved with the specific Spectral Response Function (SRF) for each EMIT band to derive band-specific solar irradiance factors. The input radiances are divided by these factors (scaled by an empirical

constant = 4.0 / 1e7 to get the scale of the radiances to be within a rough range of 0-1), effectively converting the signal into a pseudo-reflectance space.

2. We apply a cross-track radiometric correction to mitigate detector-level striping and systematic illumination artifacts across the swath. We pre-compute the means for every band at every cross-track position across all the EMIT granules. For a given pixel, the radiance is adjusted by multiplying it by the ratio of the global mean radiance for that band (averaged across all crosstrack positions) to the mean radiance for that band for the specific crosstrack ID of the pixel, thereby correcting for multiplicative striping artifacts. Figure S22 shows the difference this makes during inference.

**Model architecture.** A Swin Transformer encoder progressively downsamples the input radiance data, while a parallel convolutional decoder upsamples the features to reconstruct the tile. Skip connections link the encoder and decoder at corresponding scales to preserve multi-scale spatial information. The final decoder feature map is passed to the prediction heads (shown as "Slot linear projections") to generate multiple output slots for enhancement predictions, plume masks and plume origin masks (Figure S17).

**Ablations**

All ablations were performed on stage 1 of training over 500k steps using a batch size of 128. Post ablations, the best hyperparameters were selected before training the final model in a two stage manner as mentioned in the Methods section.

**Output scaling.** To ensure the model effectively learns from the full dynamic range of emissions, we investigated different scaling strategies designed to prioritize the accurate retrieval of lower-intensity plumes while maintaining performance on larger plumes. Both square root and log1p perform much better than no scaling (linear) but square root ended up performing marginally better.

**Plume region upweighting.** To ensure the model deals with the imbalance of plume vs non plume pixels effectively, we explore various plume region upweights for the enhancement loss and the instance segmentation losses. An upweight of 30x for plume pixels on the enhancement loss and no upweight for the instance segmentation losses worked best.

**Loss ablations.** To regularize the model's slot predictions, we added additional aggregate losses to the enhancement, plume and origin predictions. These help ensure that the amount of methane is captured correctly in a scene. We see that both the aggregate enhancement and segmentation losses help improve model performance.

**Synthetic evaluation failure cases.**

Failures primarily occur in complex scenarios where the target signal is obscured: plume overlap in high-density areas causes localization errors, while temporal intermittency and fragmented plumes lead to reduced recall (Figure S23). Additionally, low-albedo surfaces (e.g., water) result in poor signal-to-noise ratios, often masking enhancements.

**Granule level inference**

**Overview.** To facilitate global-scale methane retrieval, we extend the MAPL-EMIT framework to process contiguous EMIT granules. While the core model operates on fixed 256x256 pixel spatial windows, real-world super-emitter emissions often exceed this spatial extent. To mitigate boundary truncation, suppress duplicate detections, and seamlessly aggregate predictions across large geographical areas, we employ a strided inference and spatial clustering methodology. This also functions to create an ensemble of model predictions, giving us an additional metric of model confidence (the percentage of times a given plume appears in each prediction ensemble).

**Strided inference and spatial reconstruction.** To mitigate the artificial bisection of spatially extended plumes, the framework processes full EMIT granules using overlapping spatial strides. The model iteratively evaluates 256x256 pixel tiles with a spatial stride of 64 pixels, yielding a 75% overlap between adjacent windows. While this strided approach ensures complete morphological capture, it inherently generates redundant candidate predictions for individual physical plumes across adjacent spatial windows.

To resolve boundary artifacts and reconstruct continuous emission fields, overlapping predictions are aggregated using a spatial Hanning window (L2 distance averaging). This approach assigns higher weights to predictions near the center of each inference chunk, thereby minimizing edge-effects and seamlessly merging overlapping probability maps into a continuous global raster.

**Candidate consolidation and denoising.** While strided inference reconstructs the total spatial extent of the plumes, it produces multiple overlapping instance candidates for a single physical source. To consolidate these multiple detections, we introduce a feature-level and spatial clustering algorithm inspired by DBSCAN. Standard spatial clustering relies solely on proximity, which can erroneously merge distinct, proximal emission sources common in dense industrial regions. To preserve instance separation, our algorithm enforces two matching criteria for candidate consolidation:

1. Spatial Proximity: The Euclidean distance between the predicted plume origins must not exceed a predefined threshold of 1.5 km.
2. Enhancement Correlation: For candidate pairs meeting the proximity constraint, local enhancement patches centered on their respective origins are extracted and aligned. The Pearson correlation coefficient is computed across overlapping pixels, requiring a correlation exceeding 0.955 for the candidates to be classified as identical.

Consolidated candidates are aggregated over a joint bounding box using a regularized weighted mean. Finally, a connected components analysis is applied to the thresholded binary masks to remove high-frequency noise, discarding components smaller than 36 pixels. This yields discrete, vectorized plume polygons suitable for geospatial analysis. We filter out plumes with an inference cluster size > 13. Additionally, we provide all the plume statistics along with each plume to allow end users to dynamically select thresholds that align with their specific sensitivity and precision requirements.

**Hyperparameter optimization.** Hyperparameters were chosen in two stages. In the first stage, model training hyperparameters were tuned on synthetic granules where exact precision and recall could be calculated and optimised. In the second stage, inference hyperparameters were tuned on real-world data. This real-world data included both the EMIT L2B plumes dataset (which only allowed measuring recall as it has no negative examples) and the non-emission granules where experts expect a minimal number of plumes. The refinement focused on finding a good balance of maximizing recall on the NASA plumes while reducing the number of unexpected detections in the plume-free regions.

## Figures

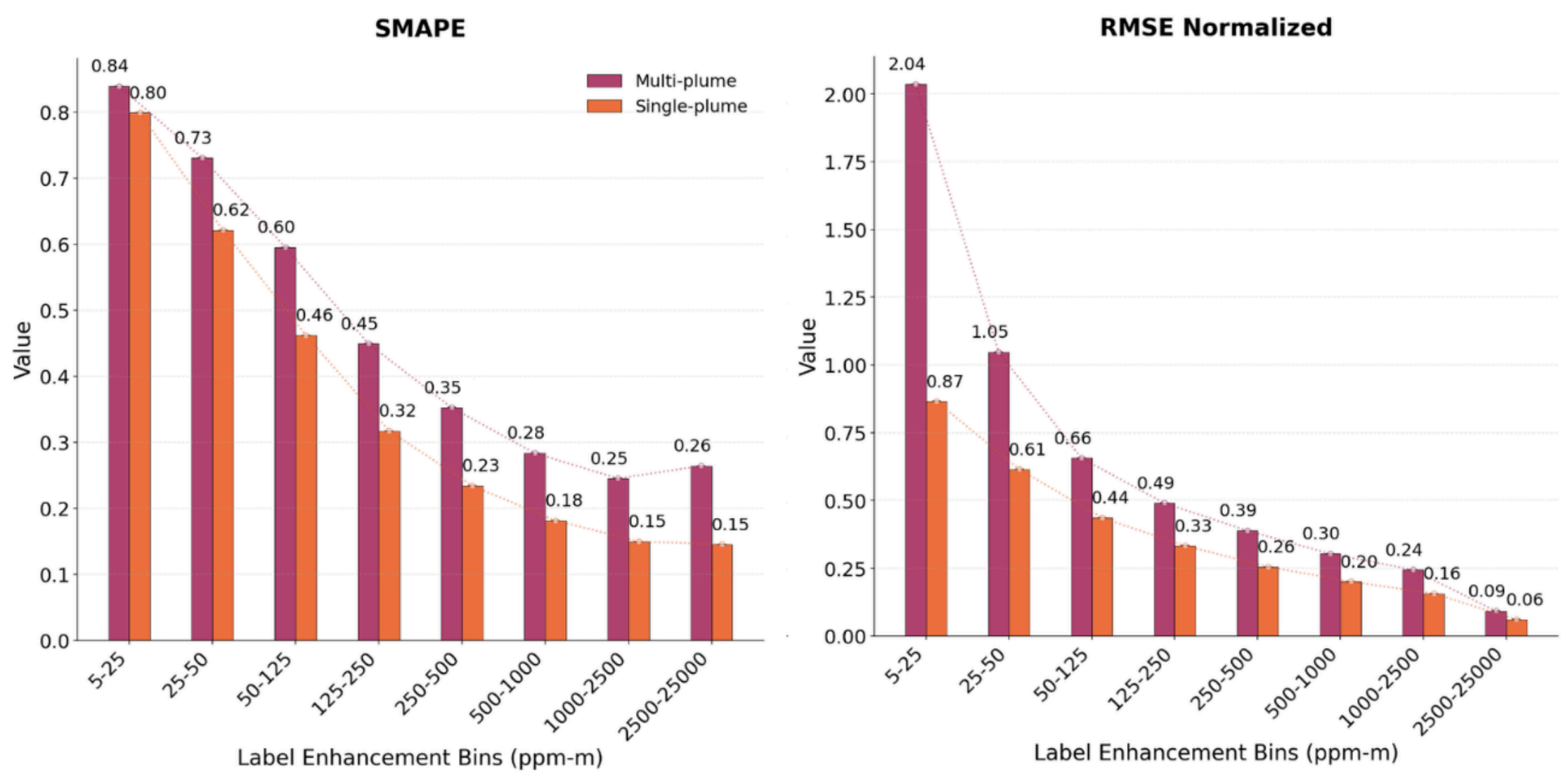

**Figure S1:** MAPL-EMIT demonstrates robust quantification of methane enhancements across a wide dynamic range. Performance metrics, including SMAPE and RMSE (normalized by mean bin enhancement), are stratified by label enhancement buckets to characterize model sensitivity from 250 to 25,000 ppm-m. Slot-level evaluation on the synthetic test split shows high accuracy from the 50-125 ppm-m bucket onwards.

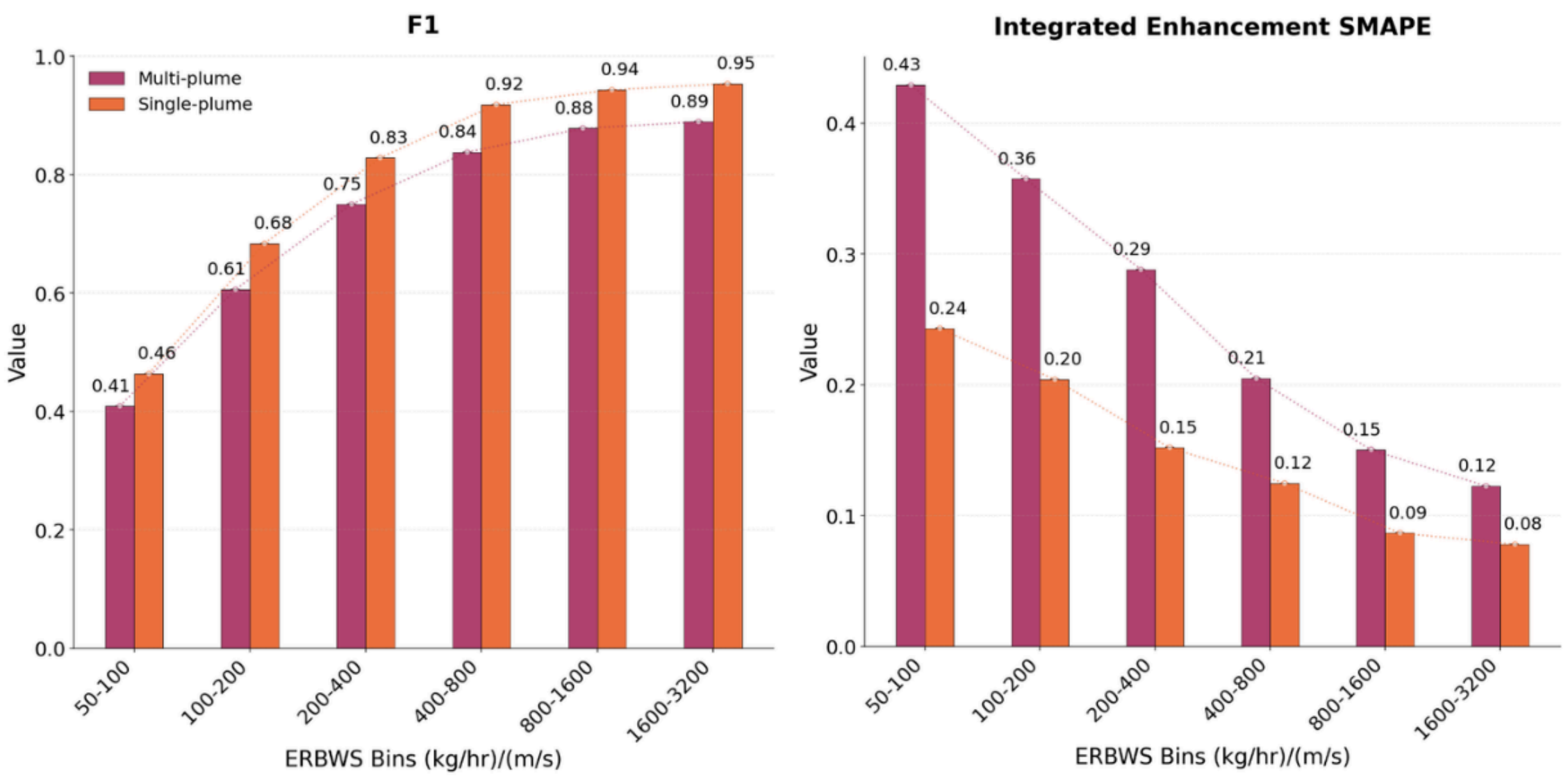

**Figure S2:** High-fidelity detection and enhancement mass quantification of MAPL-EMIT methane plumes. The figure presents the F1 score (left) and integrated enhancement SMAPE (right) stratified by ERBWS buckets. Integrated enhancement SMAPE across the synthetic test split demonstrates the model's capacity to accurately estimate the total plume mass essential for emission rate quantification.

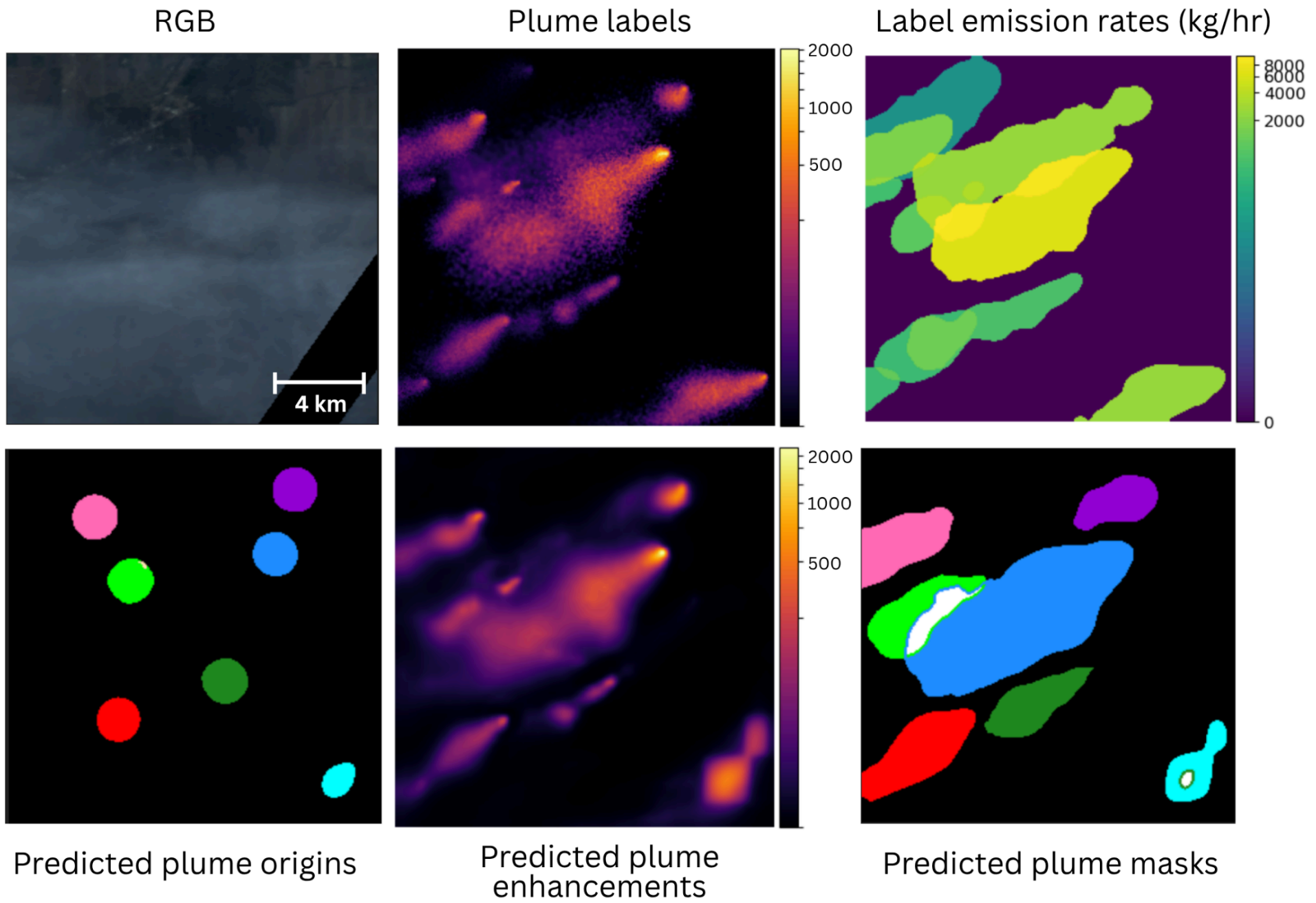

**Figure S3:** MAPL-EMIT performance on synthetic data. This example demonstrates the model's ability to disentangle overlapping plumes and accurately identify origins under varying emission rates. Each row displays (from left to right): RGB, ground-truth plume labels with mask outlines, predicted source locations, aggregated model enhancements, and predicted instance masks where overlapping segments are highlighted in white.

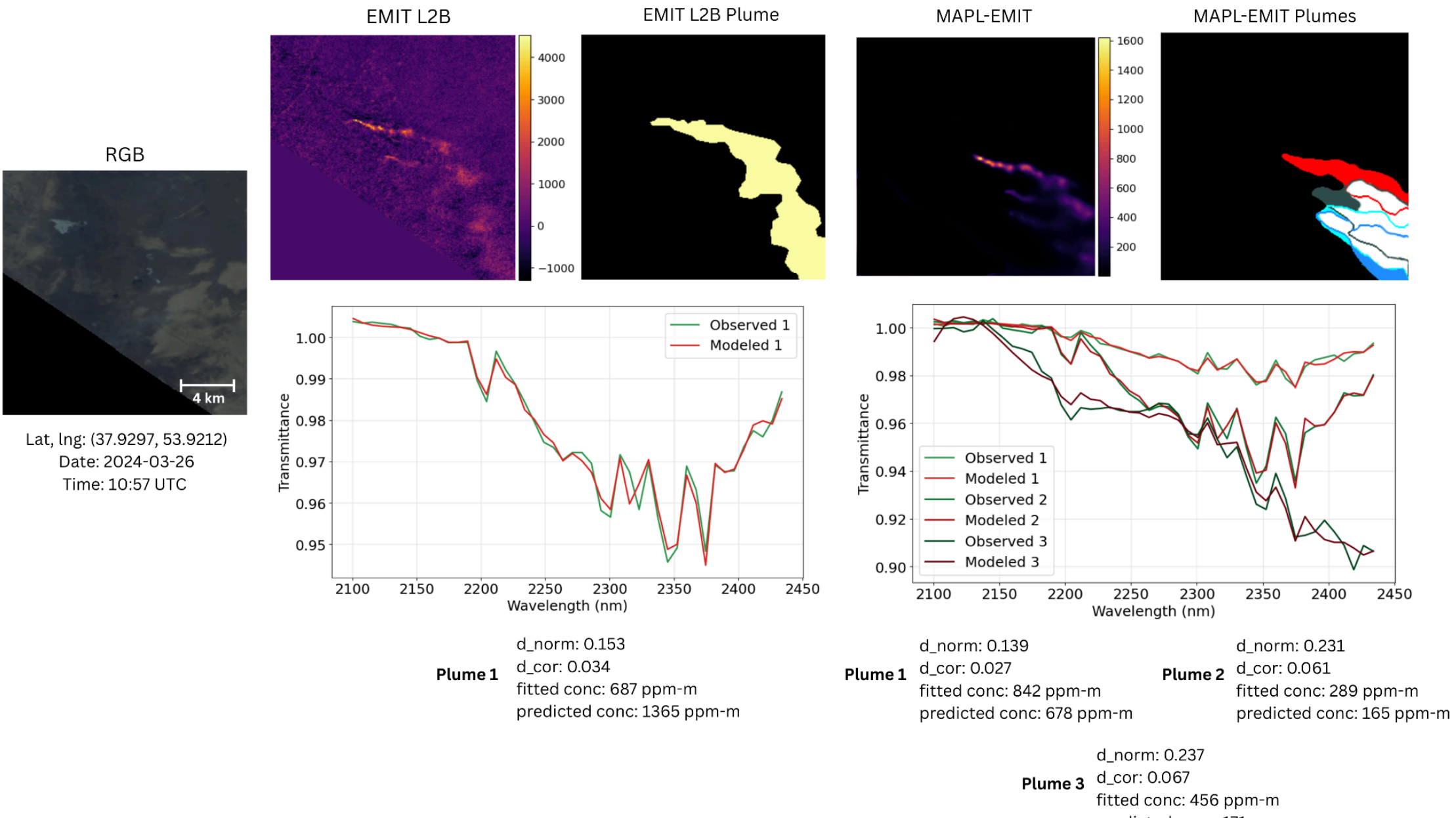

**Figure S4:** Comparative visualizations of plume detections across MAPL-EMIT predictions and EMIT L2B. From left to right: the RGB image, EMIT L2B enhancements, EMIT L2B plumes, MAPL-EMIT predicted enhancements, MAPL-EMIT instance masks, EMIT L2B spectral fit and MAPL-EMIT spectral

fits. This example illustrates the case where both MAPL-EMIT and the operational EMIT L2B identify plumes but MAPL-EMIT picks up additional plumes.

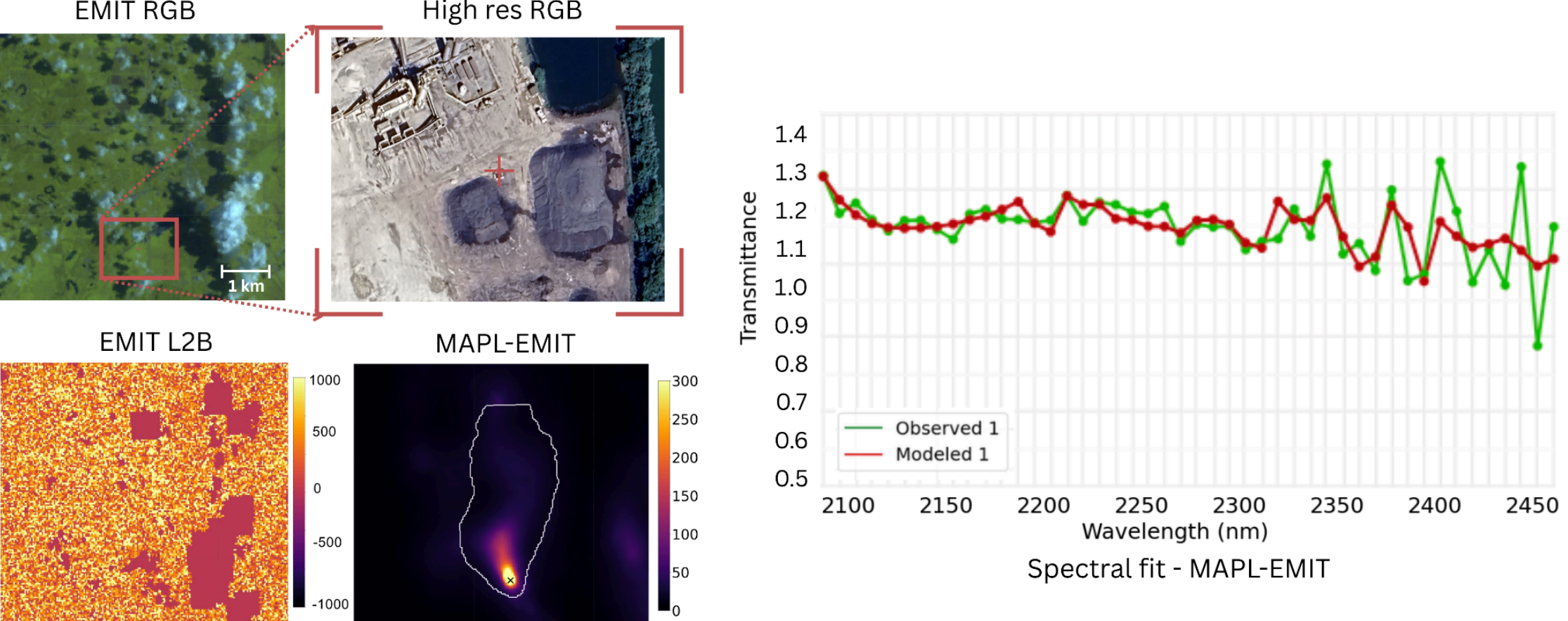

**Figure S5:** MAPL-EMIT identifies significant methane plumes that are occasionally missed by manual NASA L2B analysis. This example from a concrete/asphalt facility (29.153079, -95.436677) highlights a potential high-intensity plume detected by MAPL-EMIT that was omitted from the NASA L2B catalog, likely due to human omission or the presence of partially obstructing clouds and low-contrast enhancements in the matched-filter output.

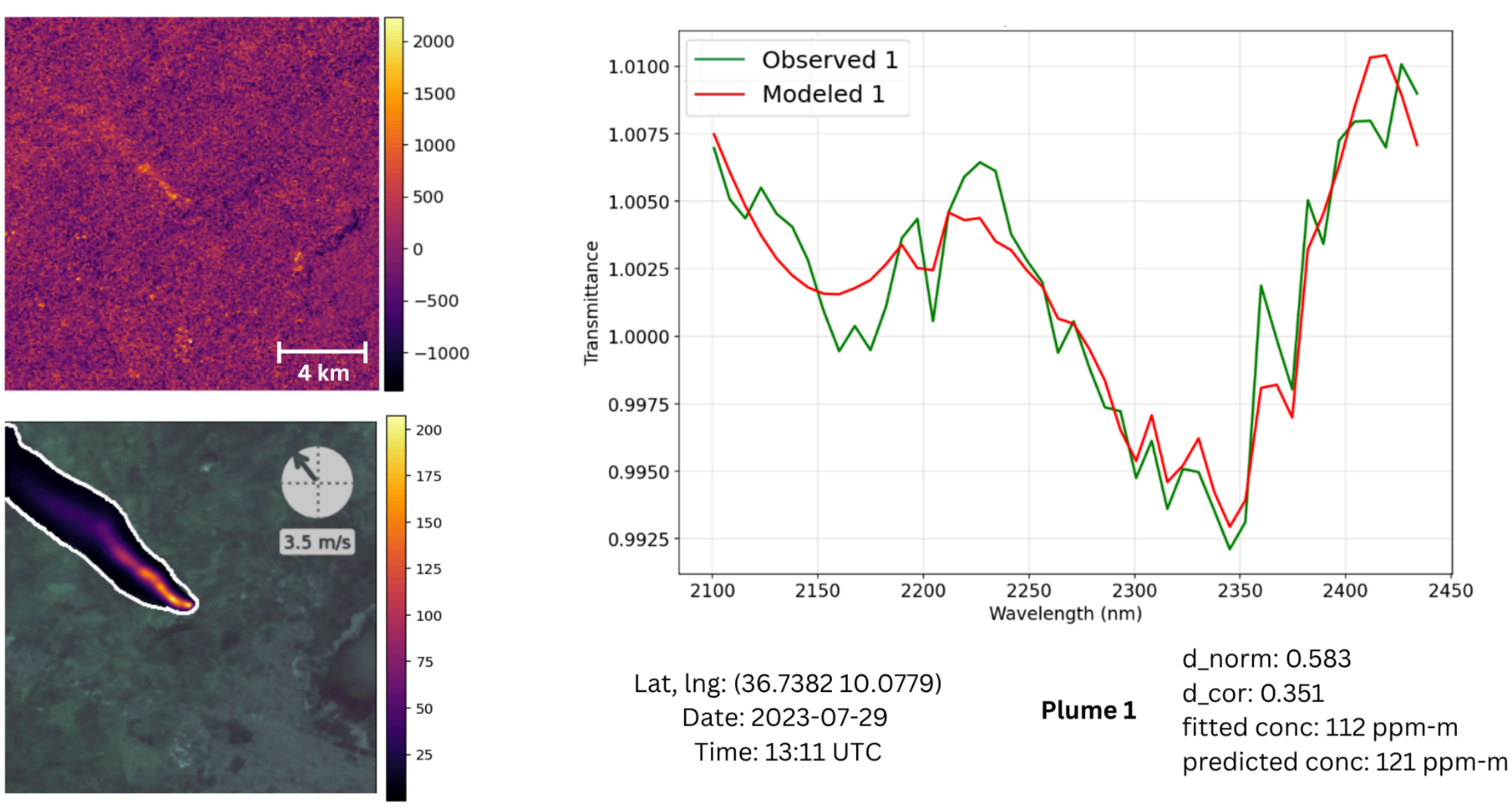

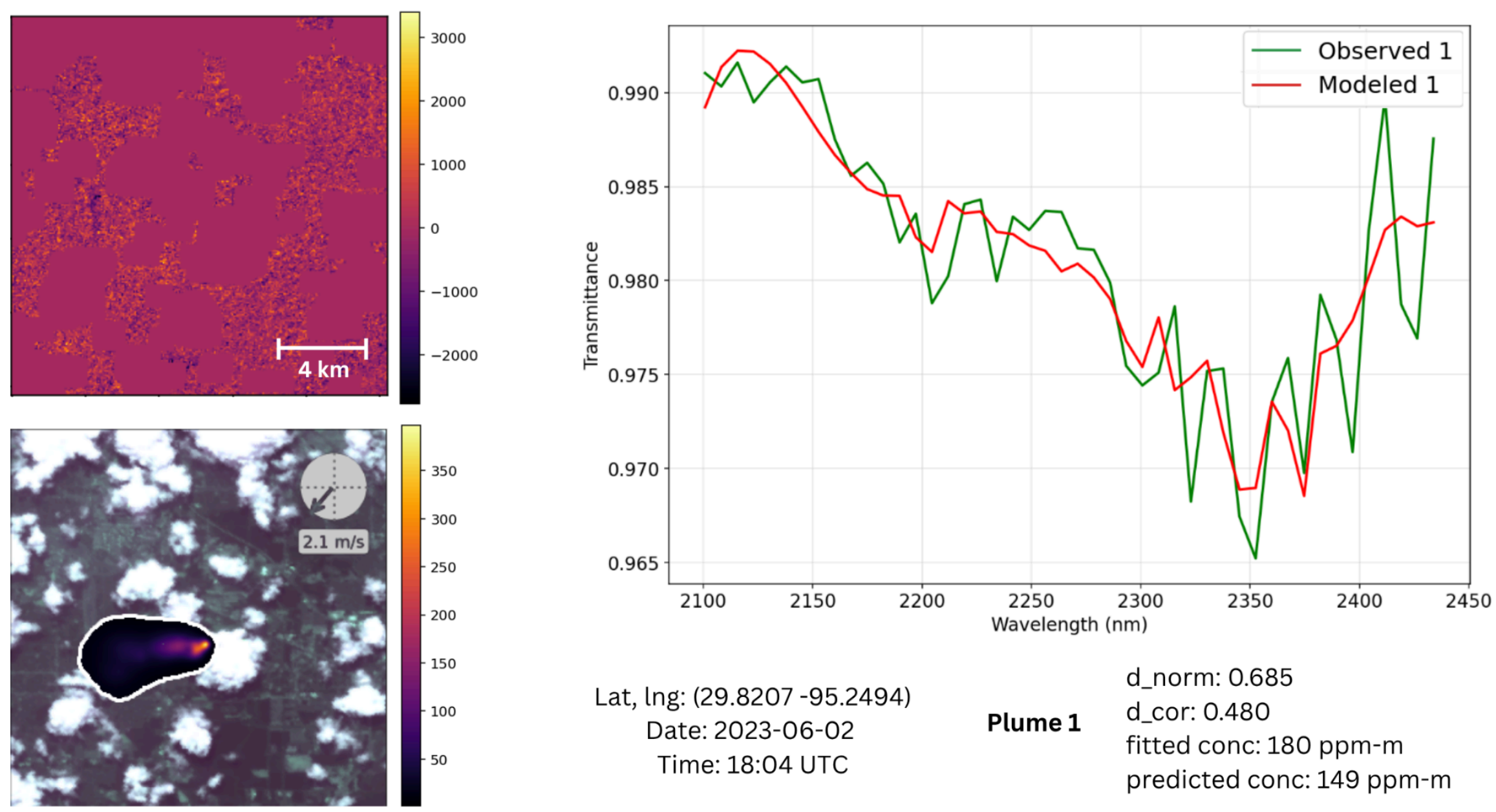

**Figure S6:** MAPL-EMIT achieves robust methane plume detection across complex and heterogeneous landfill environments globally. The panels display NASA L2B matched-filter enhancements, and the corresponding MAPL-EMIT predicted plume masks and enhancements overlaid on the EMIT RGB imagery along with a small compass on the top right showing the ERA-5 10m wind direction and the magnitude for some of the top-emitting landfills, demonstrating robust detection even under partial cloud cover.

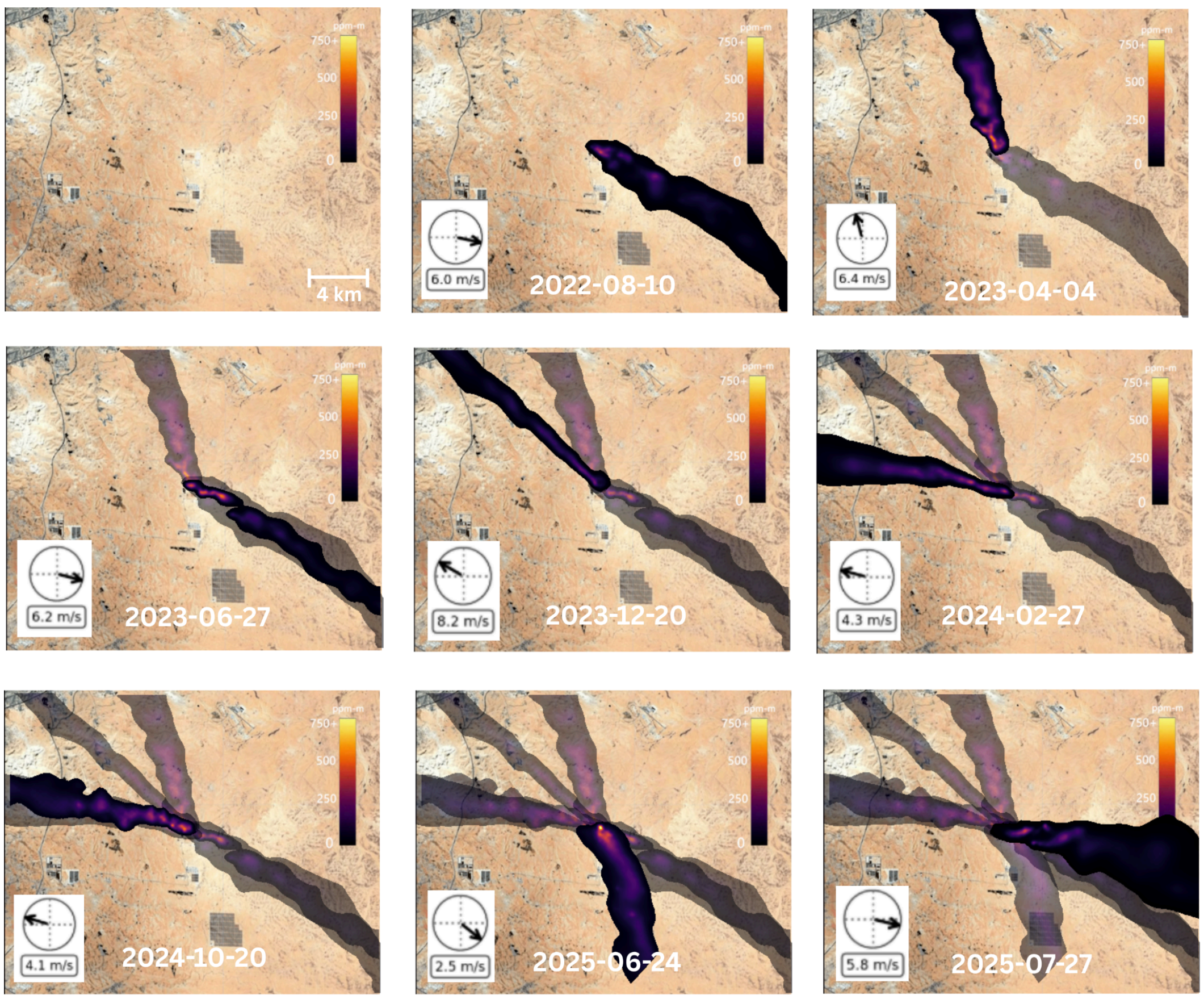

**Figure S7:** Temporal monitoring at the Amman landfill reveals persistent emission events and varying plume dispersion patterns. This time-series of detections from August 2022 to July 2025 shows active methane plumes (colored) overlaid on a cumulative history of prior detections (grey shadows), illustrating the variability in plume morphology driven by fluctuating local winds from ERA-5 10m (bottom left of each image) over a known persistent source, also indicated in the image on the top right.

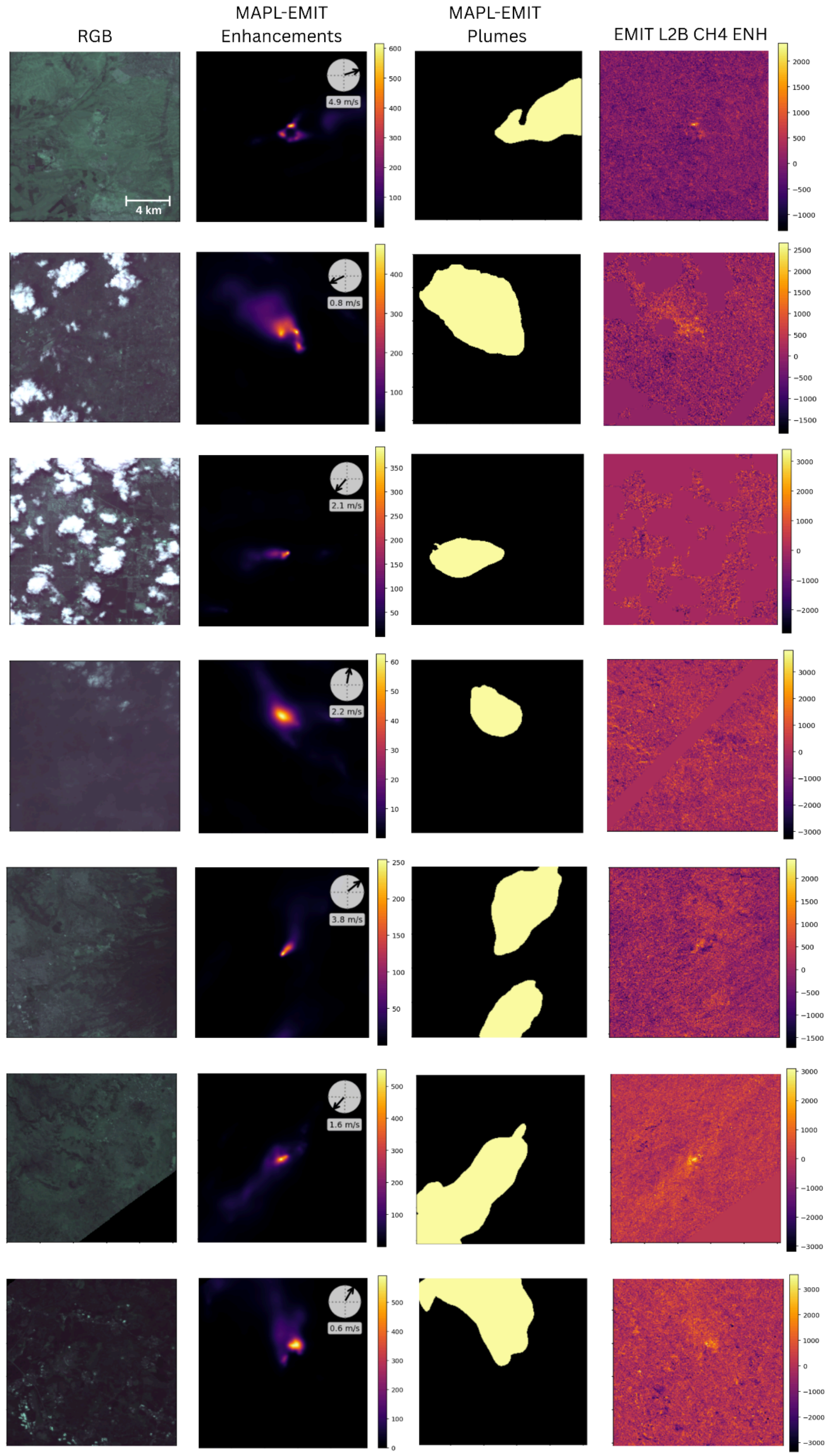

**Figure S8:** Landfill sites where methane plumes are hard to spot in EMIT L2B enhancements but are clearly identified by MAPL-EMIT. Each row corresponds to one location. The MAPL-EMIT enhancement contains the wind direction on the top right as well. For samples 1, 6, and 7, the plume extent and direction are difficult to discern from L2B enhancements; in samples 2, 3, and 4, identification is hindered

by masked regions in the standard L2B data; and in sample 5, the plume is not visible in the L2B enhancement due to a low SNR.

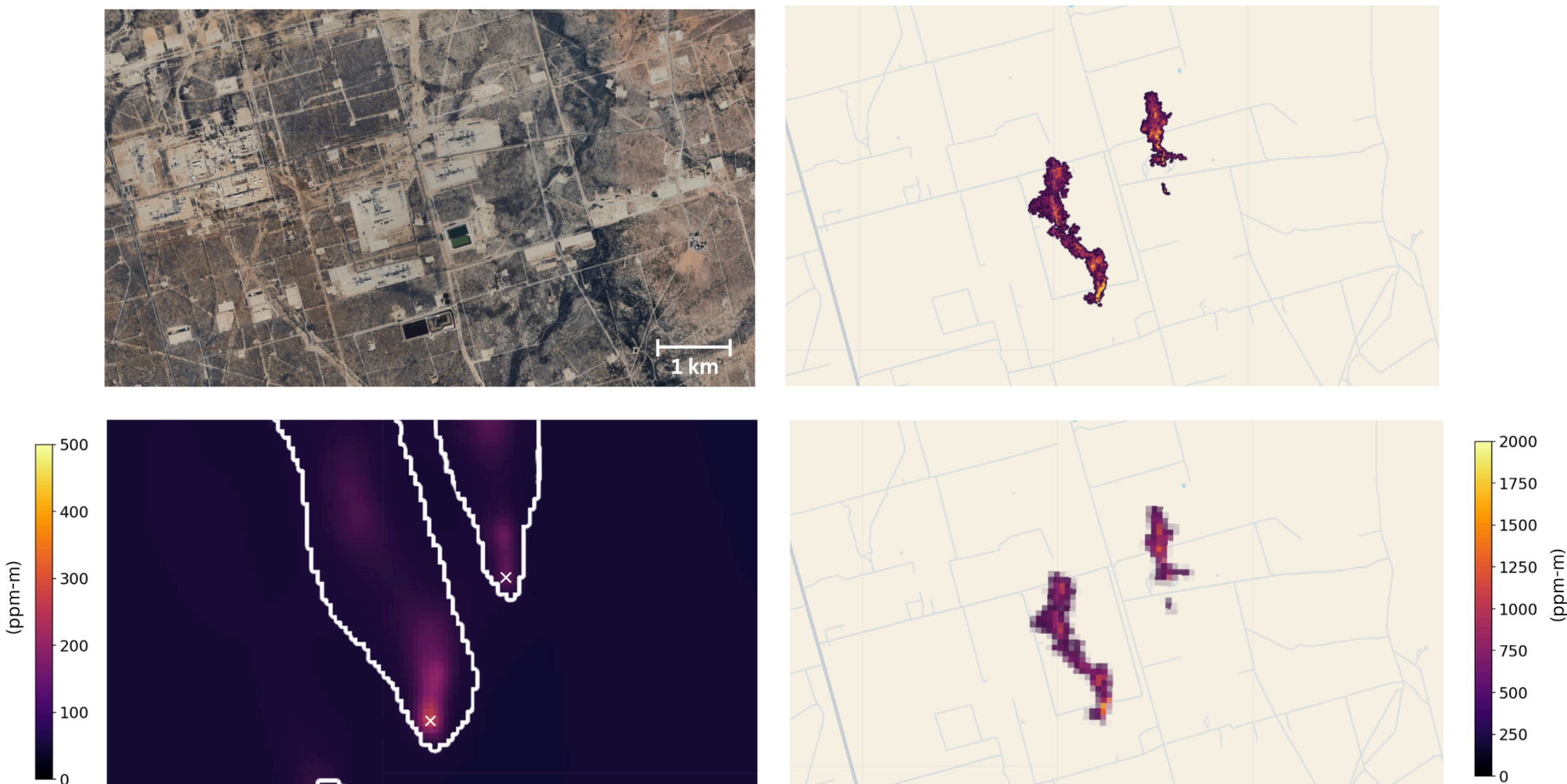

**Figure S9:** Coincident airborne observations from the Global Airborne Observatory (GAO) validate the accuracy of EMIT-based detections. The panels show high res RGB imagery, GAO full resolution plumes, MAPL-EMIT enhancements with plume outlines and GAO plumes resampled to 60m spatial resolution to match EMIT. Qualitative verification of EMIT-based detections against coincident GAO airborne data at multiple locations shows the model's ability to accurately reconstruct plume morphology and capture enhancements confirmed by the higher-SNR aerial data.

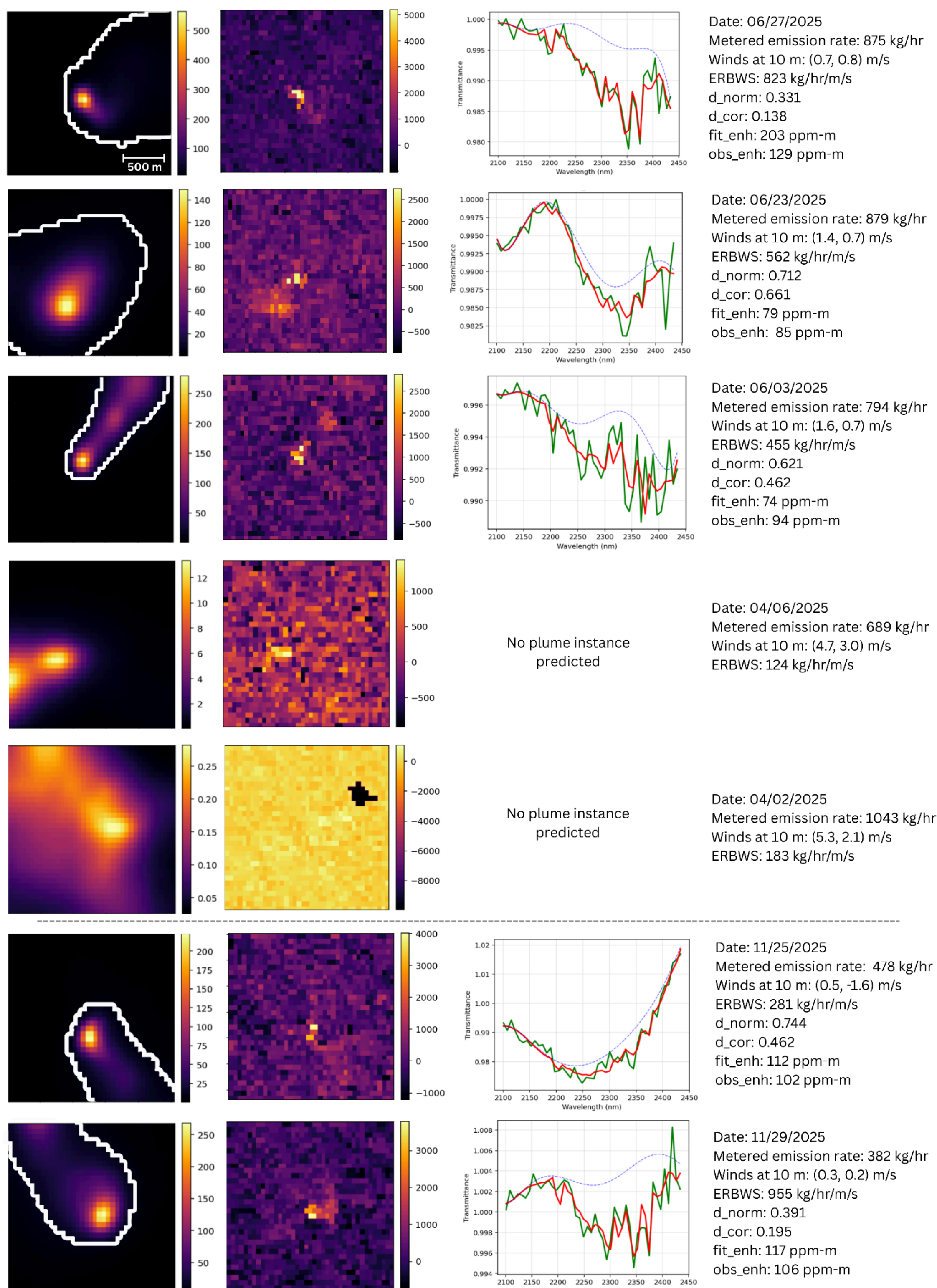

**Figure S10:** Controlled release experiments validate MAPL-EMIT's detection sensitivity and physical consistency. The figure compares MAPL-EMIT enhancements and plume masks against NASA L2B enhancements for seven controlled methane releases, showing successful detection of medium-intensity plumes with physical validity confirmed by independent spectral fit metrics.

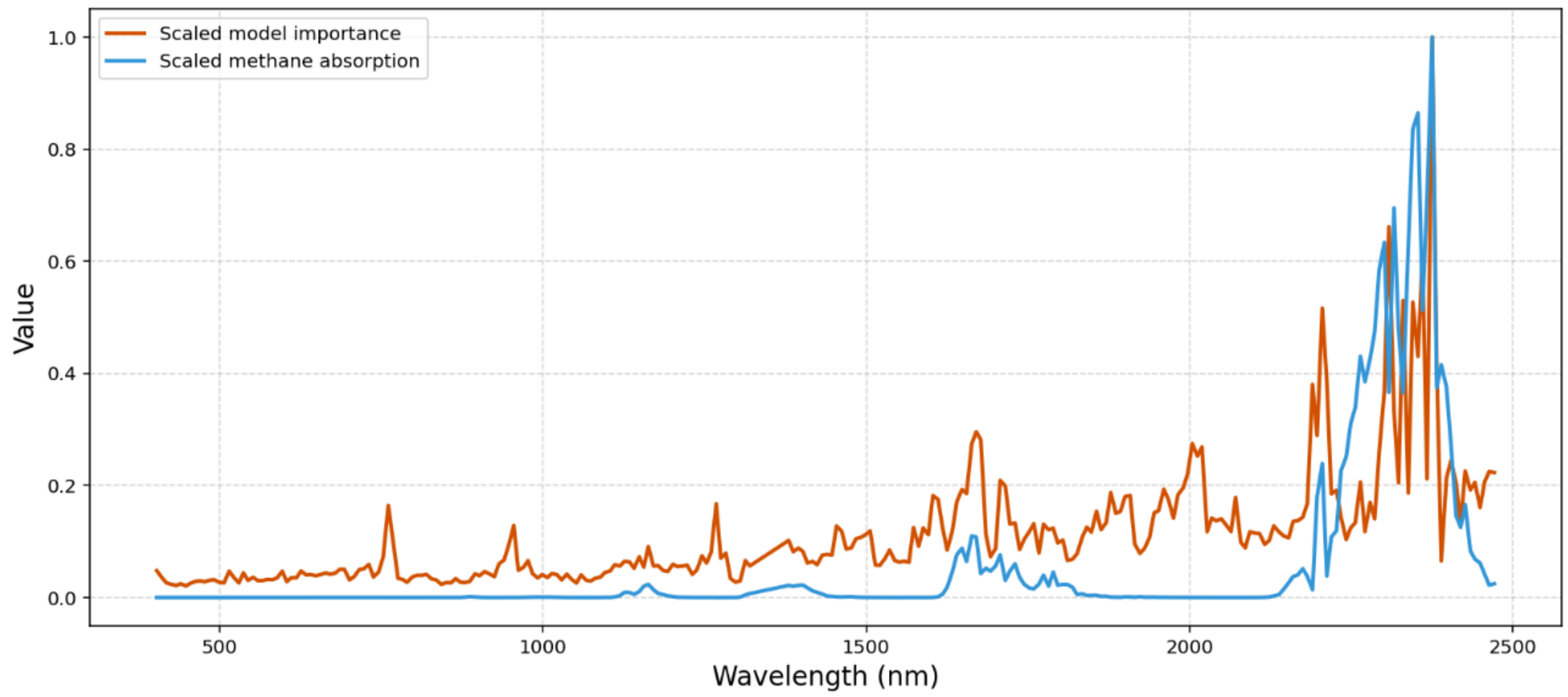

**Figure S11:** The deep learning model prioritizes physically relevant spectral bands associated with methane absorption manifolds. A comparison between learned patch embedding weights and the methane absorption curve shows that the model's attention peaks align with the three primary methane absorption features in the SWIR, confirming it effectively leverages relevant physical signals.

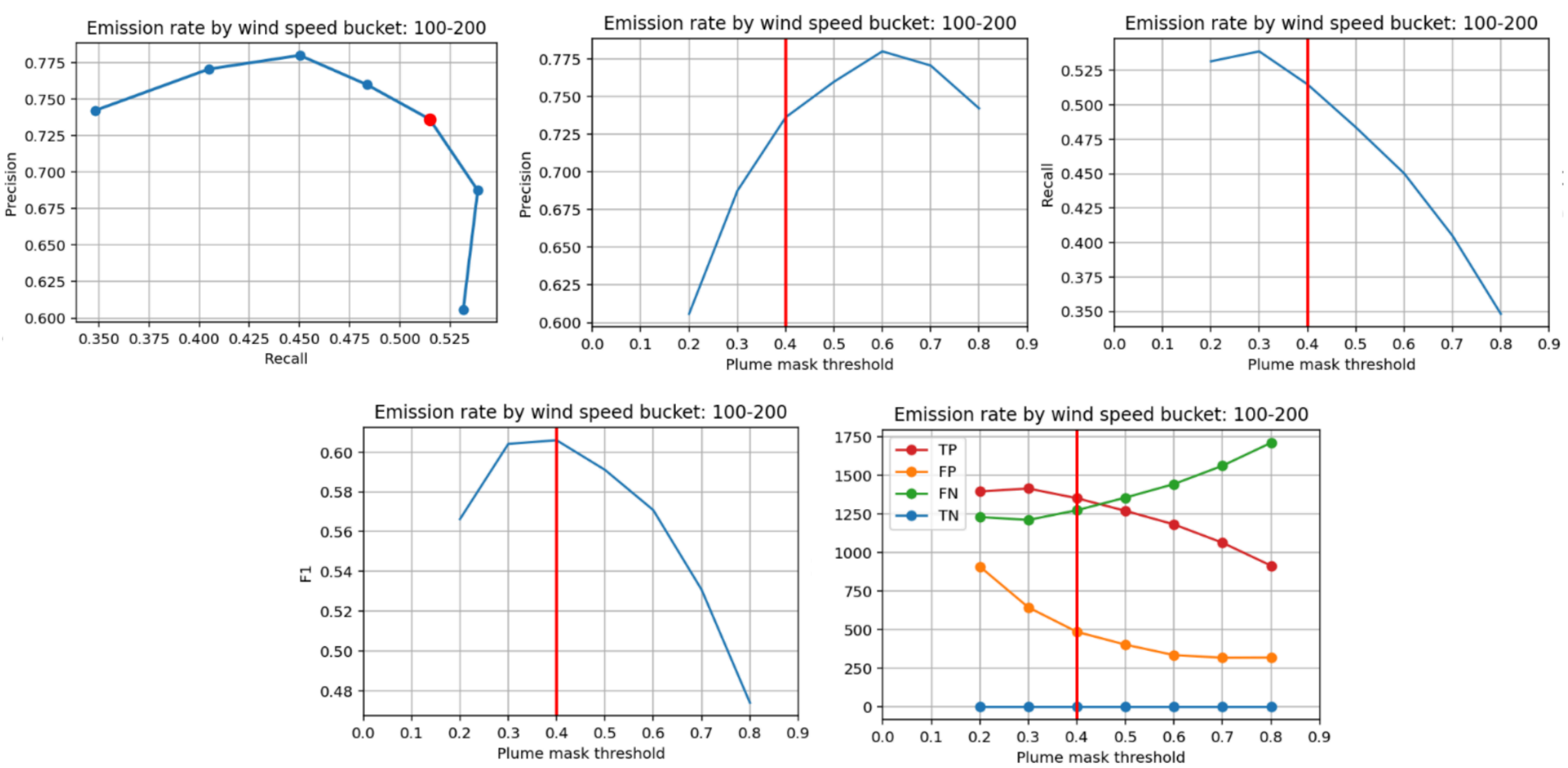

**Figure S12:** Precision and recall trade-offs can be optimized by adjusting the model's plume probability threshold. The top row displays the Precision-Recall (PR) curve and the relationship between classification outcomes across varying thresholds, while the bottom row shows F1, precision, and recall for the 100-200 ERBWS bucket, justifying the selection of an operational threshold.

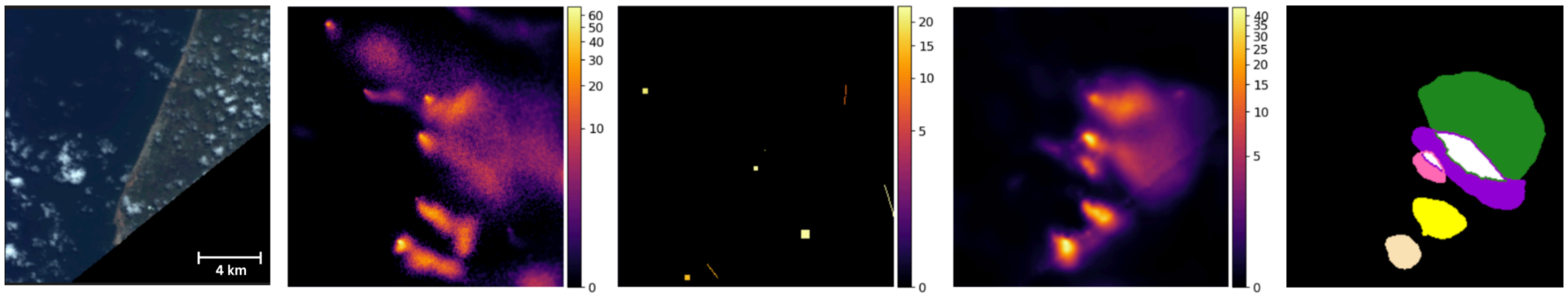

**Figure S13:** MAPL-EMIT effectively discriminates genuine methane plumes from non-physical surface artifacts. The panels display (from left to right): RGB context imagery, ground-truth plume labels, simulated surface artifacts designed to mimic methane's spectral signature but lacking realistic plume morphology, model-predicted methane enhancements, and predicted plume masks. The model successfully suppresses these non-physical spectral anomalies, accurately identifying genuine plumes while ignoring the simulated surface noise.

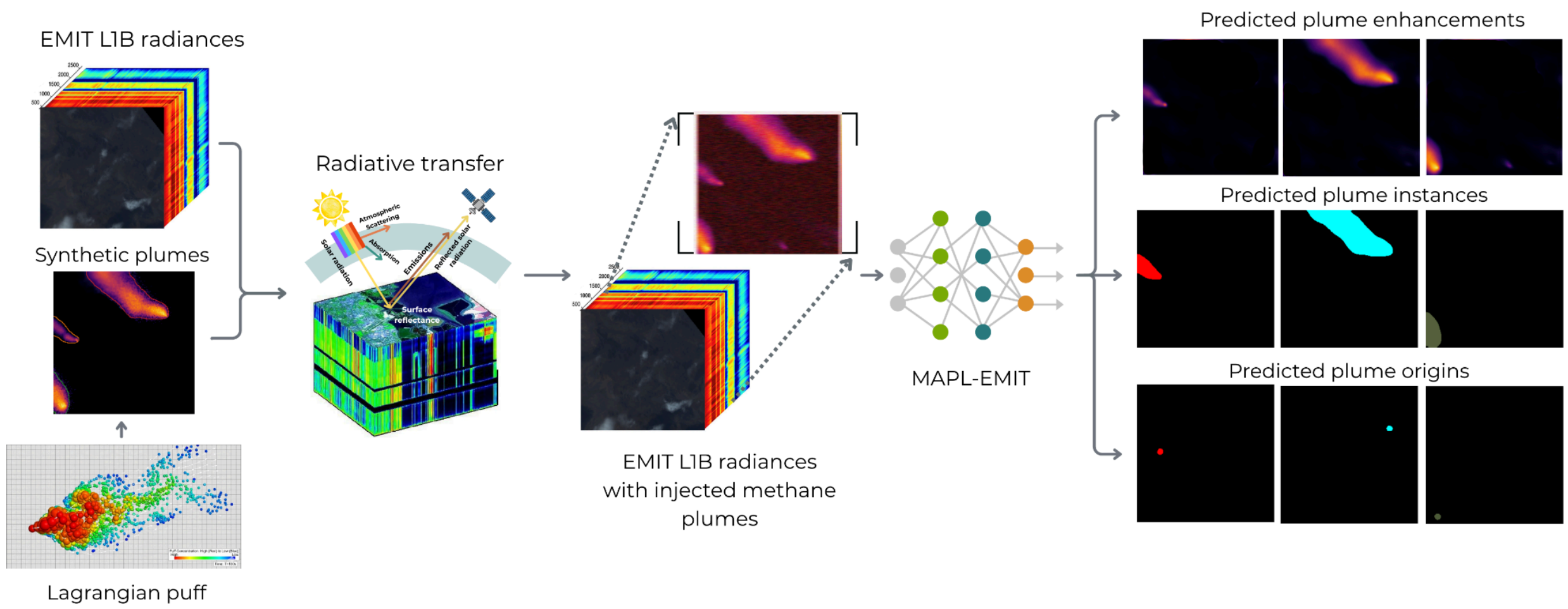

**Figure S14:** The MAPL-EMIT model utilizes a comprehensive synthetic-to-real training pipeline to overcome data scarcity. This methodological overview illustrates the integration of EMIT L1B radiances with synthetic plumes generated via a Lagrangian puff model and physics-based radiative transfer, forming the basis for training the joint prediction heads.

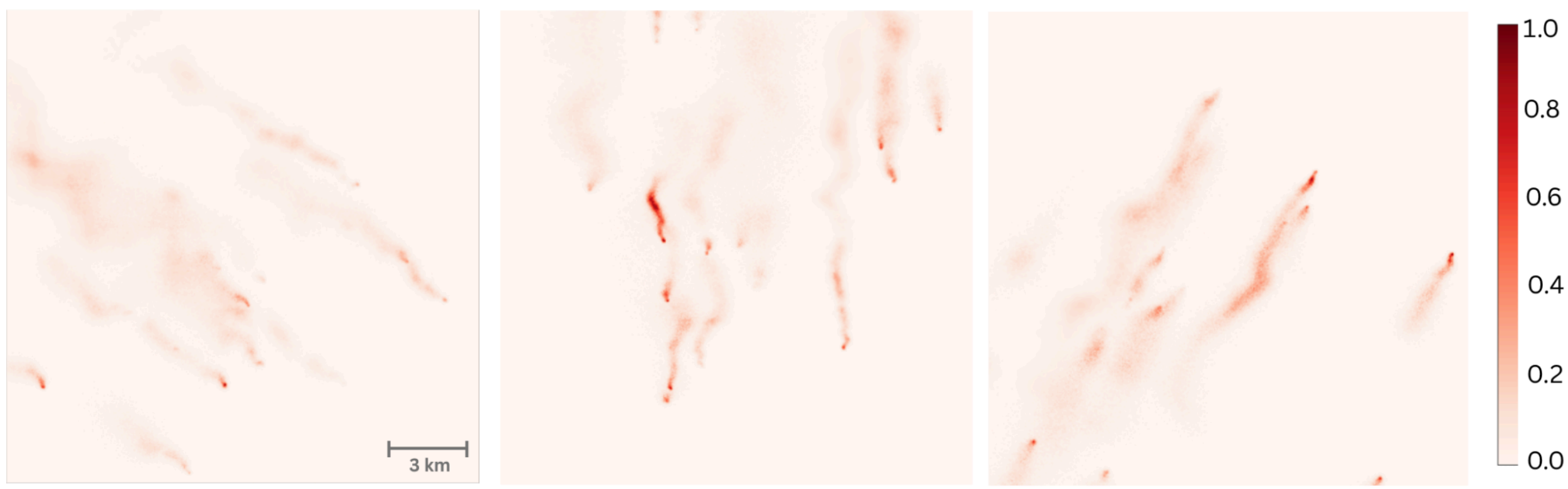

**Figure S15:** Lagrangian puff simulations capture the complex turbulence and intermittency of real-world methane release events. Examples of synthetic 60m-resolution plume tiles show the range of morphologies and intensities used during training, providing the high-fidelity ground truth necessary for the model to learn spatial and spectral gas features.

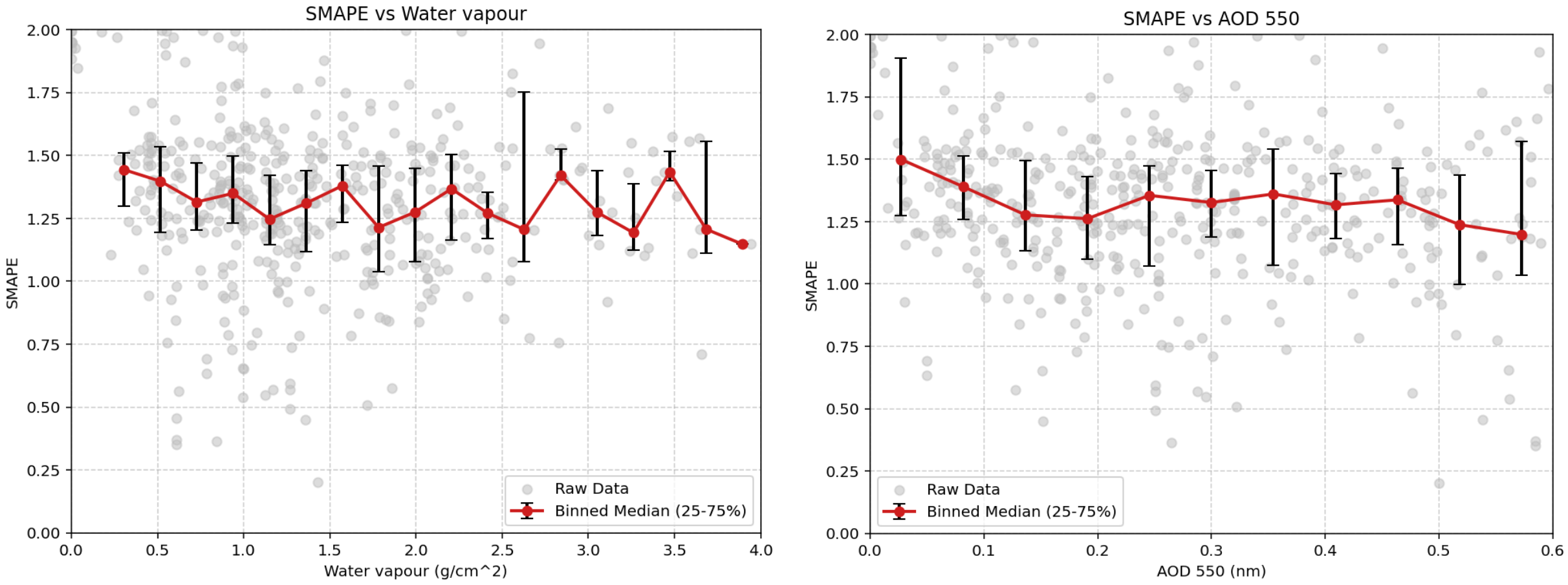

**Figure S16:** MAPL-EMIT methane retrievals remain robust across a wide range of water vapor and aerosol concentrations. Plots of integrated enhancement SMAPE versus water vapor column density (left) and Aerosol Optical Depth (right) on the NASA plumes dataset show no significant correlation between retrieval error and these atmospheric variables.

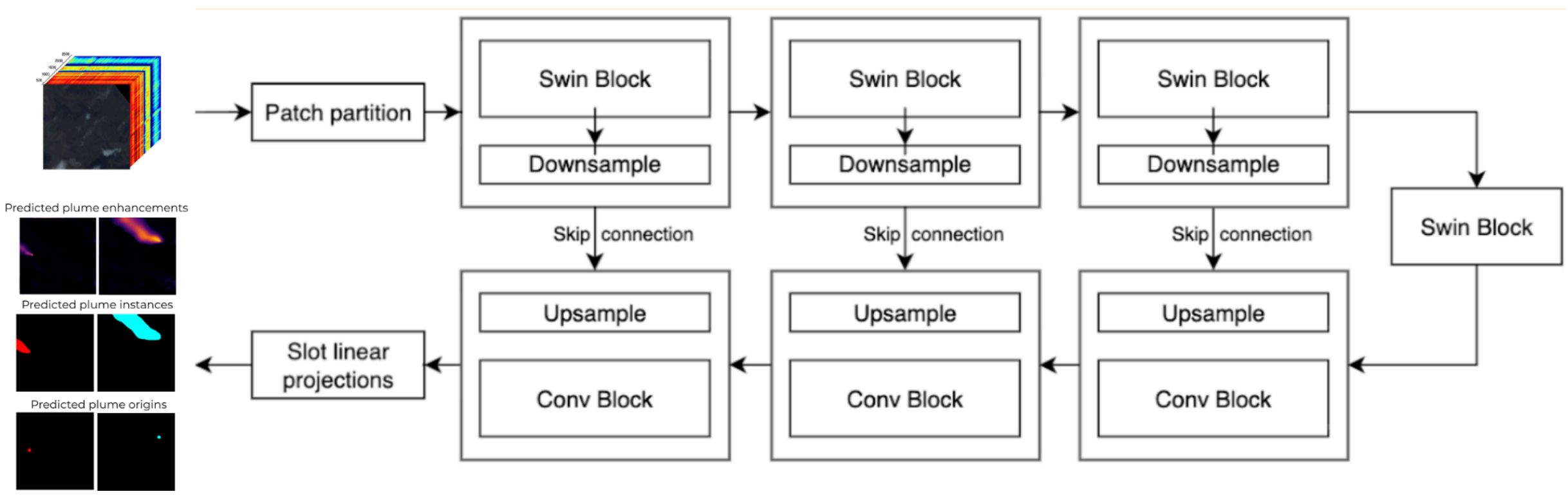

**Figure S17:** A U-Net-like architecture with a Swin Transformer encoder enables multi-scale feature extraction for methane retrieval. The diagram details the model structure, featuring a Swin-v2-S encoder for feature extraction from hyperspectral inputs and a convolutional decoder with skip connections that output parallel predictions for enhancements, masks, and origins.

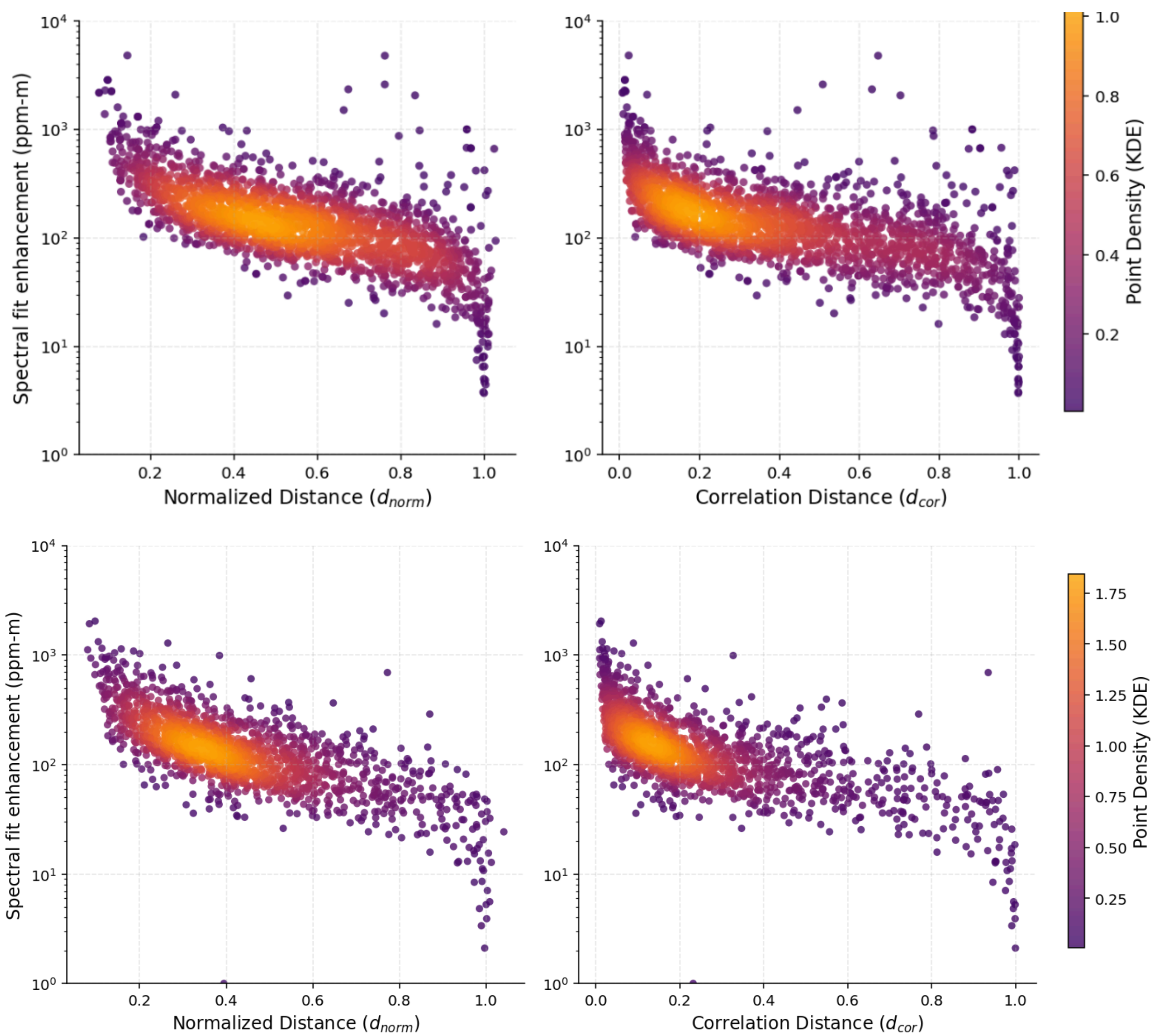

**Figure S18:** Spectral fit metrics provide independent physical verification of real-world methane plume detections. Distributions of normalized distance (d_norm) and correlation distance (d_cor) for MAPL-EMIT (top) and NASA L2B (bottom) across identical granules show that the model's detections consistently align with expected methane absorption manifolds.

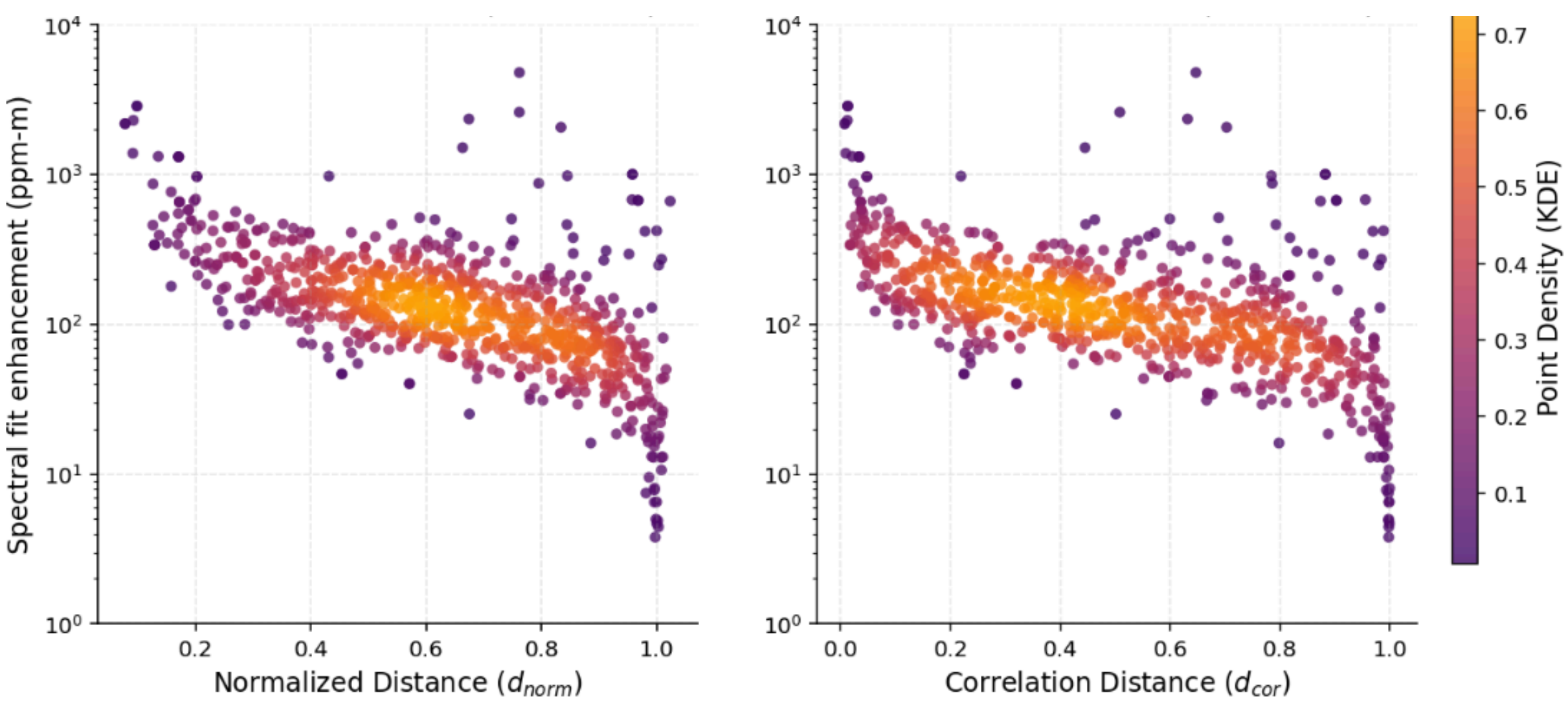

**Figure S19:** MAPL-EMIT identifies hundreds of plausible methane plumes missed by established manual catalogs. Spectral fit analysis of plumes detected by MAPL-EMIT but absent from the NASA L2B catalog indicates that approximately 67% exhibit spectral signatures consistent with genuine methane emissions.

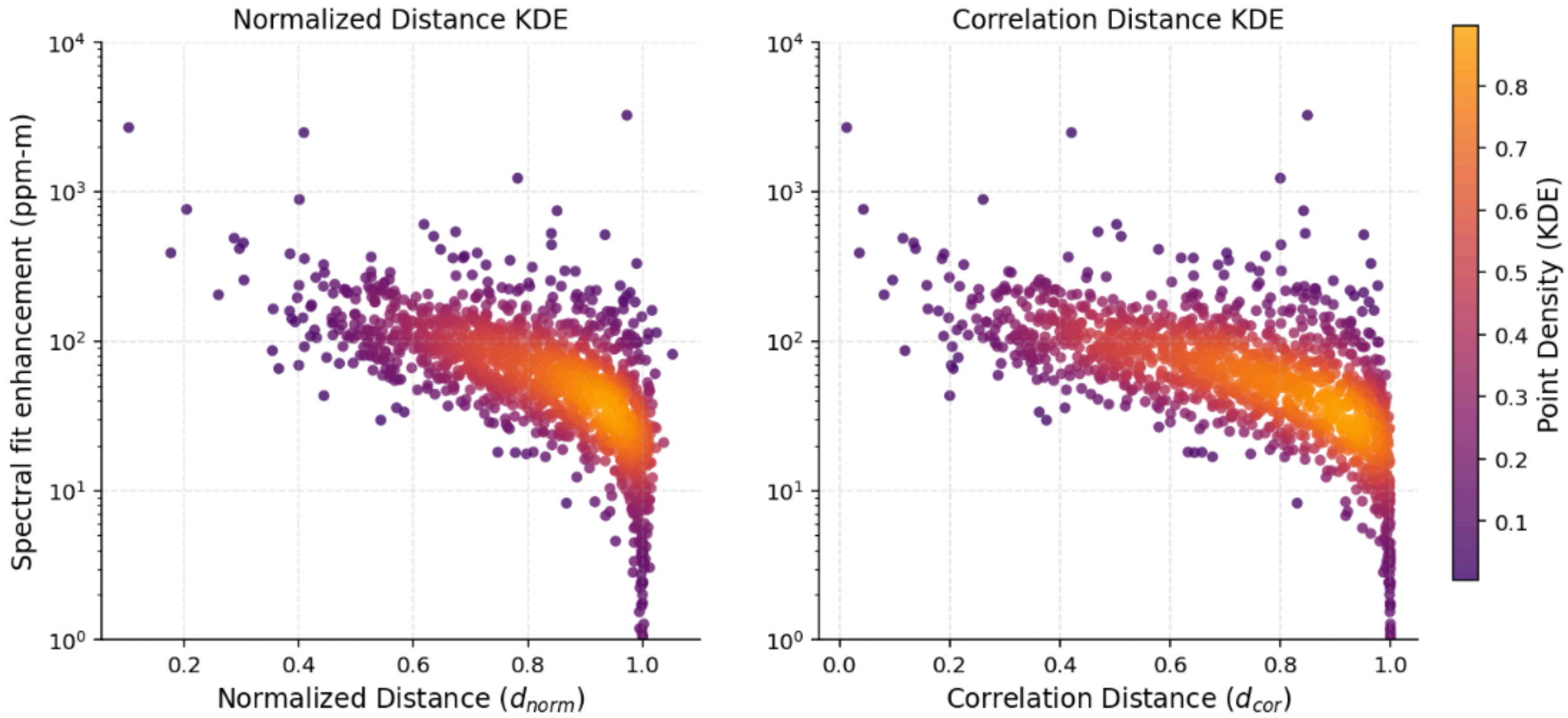

**Figure S20:** Evaluation on granules with no identified plumes from NASA provides a conservative estimate of the model's false-positive rate. Analysis of 1,513 detections in 20,000 granules expected to be plume-free shows that 355 are potentially real emissions based on spectral fit metrics, with the remainder defining the limit of retrieval artifacts.

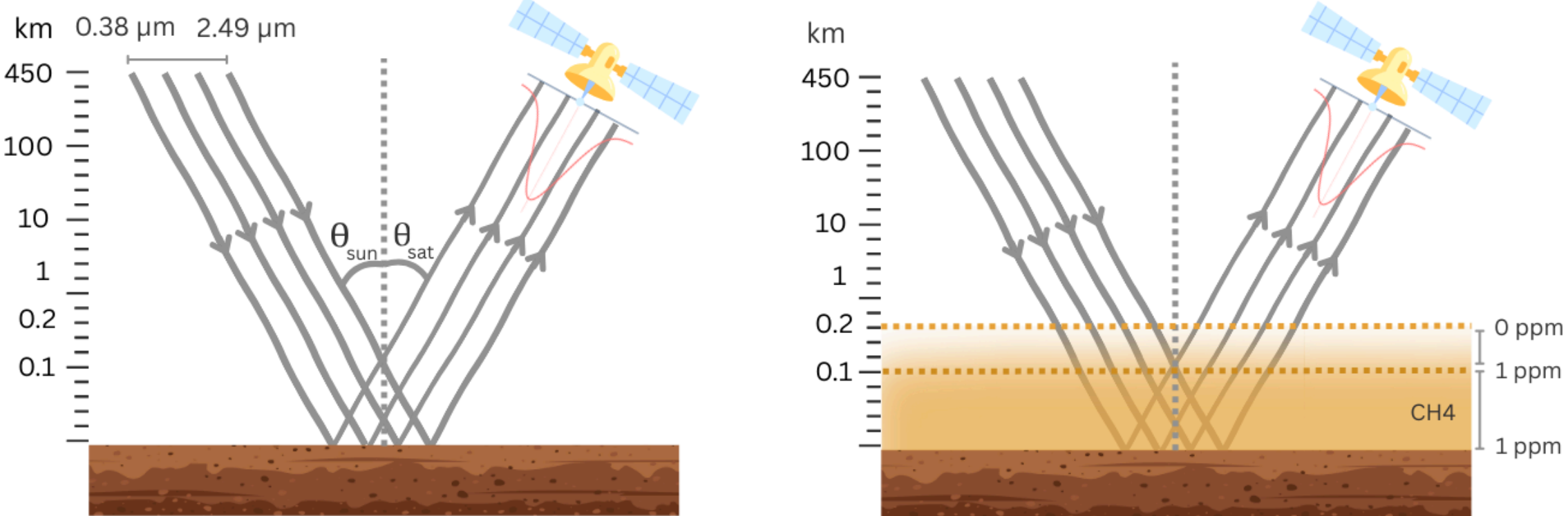

**Figure S21:** Radiative transfer modeling accounts for column-integrated absorption along the solar and satellite photon paths. The illustration depicts the simplified Beer-Lambert absorption model used to calculate transmittances for background and methane-enhanced atmospheres, accounting for total path length as a function of viewing geometry.

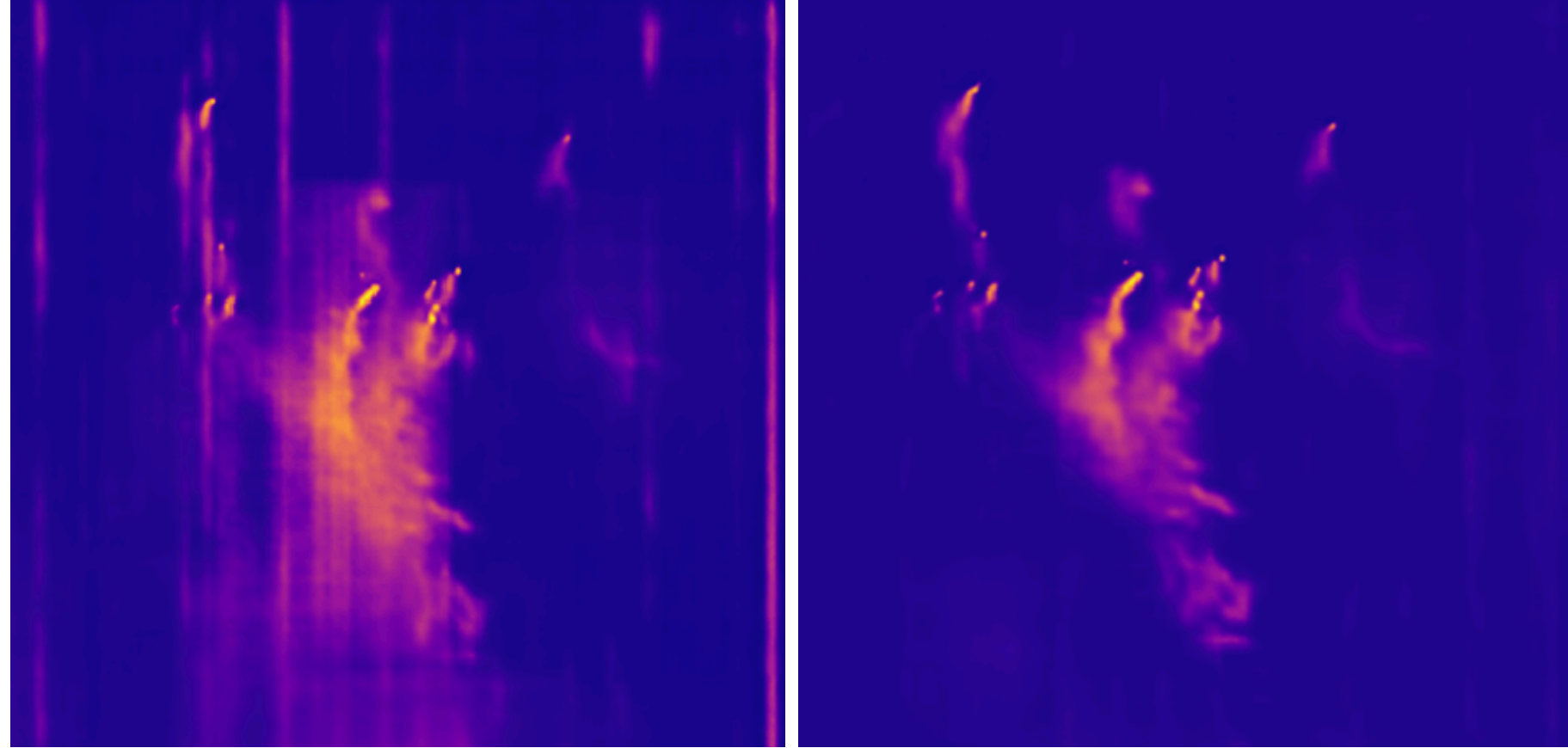

**Figure S22:** Cross-track radiometric correction effectively mitigates detector-level striping artifacts in EMIT granules. A comparison of model outputs with and without cross-track normalization demonstrates that correcting for systematic illumination differences across the sensor swath prevents the hallucination of features tied to detector geometry.

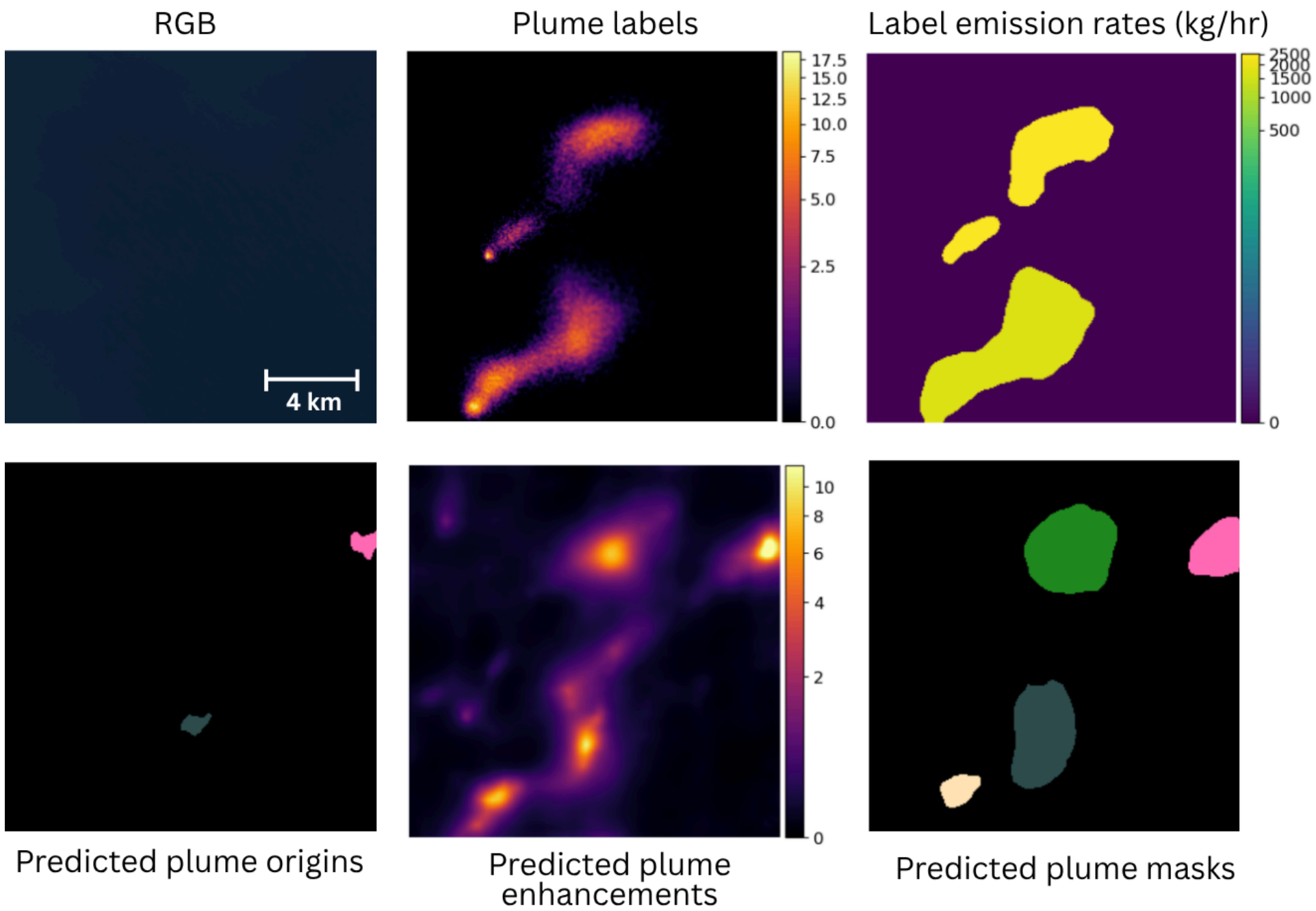

**Figure S23:** Identification of characteristic failure modes guides future improvements to the deep learning framework. Representative examples of retrieval errors include fragmented masks for highly intermittent plumes over low-reflectance terrain and challenges in perfectly delineating plumes in scenarios with extreme spatial overlap.

**Tables**

**Table S1:** Hyperparameters used for plume simulation. This table lists the key parameters and their corresponding values utilized in the generation of the synthetic plume dataset.

| **Model** | **H-param** | **Description** | **Value / Uniformly sampled from** |
|---|---|---|---|
| Plume model | Num plumes | The number of plumes in a tile | (1, 10) |
| | Simulation duration | The duration of the total simulation | 3000 seconds |
| | Plume duration | The duration each plume is emitting for. Note that every plume ends emitting at the end of the simulation itself, so the start of the plume emission is set to Simulation duration - Plume duration | Uniformly pick a bucket and then uniformly sample from the bucket. Buckets:<br>(250, 550),<br>(550, 1200),<br>(1200, 2400),<br>(2400, 3000) |
| | Simulation region | The size of the simulation region. This region is then cropped (random crop during training, center crop during eval) to 256x256 pixels which is the size of each EMIT tile. | 348x348 pixels at 60m resolution. Total region size = 23,040 |
| | Center relative diff scale | Each puff moves with x' = x + (u + rand * center_relative_diff_scale) *dt, this adds noise into the wind field for each puff at every timestep | (10, 20) |
| | Puff initial radius | The initial radius of the puff spawned in m | (3, 20) |
| | Puff spread rate | The puff's radius increases by pow(puff_spread_rate, dt) every timestep | (1.002, 1.003) |
| Wind model | Mean wind speed | Initial mean wind speed in m/s across the wind field | (0, 10) |
| | Velocity clip | Maximum wind speed across the wind field across time | 30 |
| | Diffusivity range | Diffusivity constant | (30, 50) |
| | Puff release rate | Number of puffs released for each plume from the origin | Mean = 20, std = 4 |

**Table S2:** Impact of output scaling on retrieval performance. Analysis of scaling methods conducted on the synthetic multi-plume validation split, focusing on the 100–200 ERBWS stratification bucket. Note that the ablation was performed on stage 1 of training (500k steps with 128 batch size).

| **Output scaling** | **F1** | **Precision** | **Recall** |
|---|---|---|---|
| **Sqrt (Baseline)** | **0.609** | **0.763** | 0.507 |
| Linear | 0.569 | 0.712 | 0.474 |
| Log1p | 0.605 | 0.746 | **0.508** |

**Table S3:** Impact of removing aggregate losses on retrieval performance. Analysis was conducted on the synthetic multi-plume validation split, focusing on the 100–200 ERBWS stratification bucket. Note that the ablation was performed on stage 1 of training (500k steps with 128 batch size).

| **Experiment** | **F1** | **Precision** | **Recall** |
|---|---|---|---|
| **Baseline** | **0.609** | **0.763** | **0.507** |
| No aggregate regression loss | 0.589 | 0.757 | 0.482 |
| No aggregate segmentation losses | 0.572 | 0.742 | 0.465 |
| No aggregate losses | 0.555 | 0.744 | 0.442 |

**Table S4:** Impact of plume upweight (enhancement, semantic) on retrieval performance. Analysis was conducted on the synthetic multi-plume validation split, focusing on the 100–200 ERBWS stratification bucket. Note that the ablation was performed on stage 1 of training (500k steps with 128 batch size).

| **Experiment** | **F1** | **Precision** | **Recall** |
|---|---|---|---|
| **30x, 1x (Baseline)** | **0.609** | **0.763** | 0.507 |
| 1x, 1x | 0.567 | 0.735 | 0.461 |
| 15x, 1x | 0.599 | 0.754 | 0.496 |
| 50x, 1x | 0.604 | 0.745 | 0.507 |
| 1x, 2x | 0.569 | 0.692 | 0.483 |
| 15x, 2x | 0.600 | 0.714 | **0.518** |
| 30x, 2x | 0.584 | 0.705 | 0.499 |
| 50x, 2x | 0.598 | 0.721 | 0.510 |

**SI References**